\documentclass[a4paper]{article}

\usepackage[numbers]{natbib}

\usepackage{authblk}

\usepackage{todonotes}
\usepackage{mathtools}
\newcommand{\nd}{\noindent}

\usepackage{booktabs}

\usepackage{array} % <-- Required to define custom columns
\newcolumntype{L}{>{\raggedright\arraybackslash}p{0.2\textwidth}}

\usepackage{graphicx}
\usepackage{tabularx}

\usepackage[table]{xcolor}
\newcommand{\heat}[2]{%
  \ifnum#1<50
    \cellcolor{cyan!\the\numexpr#1*2\relax!white}\textcolor{black}{#2}%
  \else
    \cellcolor{blue!\the\numexpr(#1-50)*2\relax!cyan}%
    \ifnum#1>70\textcolor{white}{#2}\else\textcolor{black}{#2}\fi
  \fi
}

\title{A Fuzzy Rule-based Neuro-Symbolic Approach for Pipe Severity Prediction in Sewer Networks}

\author[2]{Ngoc Thai Le}
\author[2]{Thanh Ma}
\author[1]{Umberto Straccia\thanks{Corresponding author}}

\affil[1]{Istituto di Scienza e Tecnologie dell'Informazione, CNR - ISTI, Pisa, Italy}
\affil[2]{Can Tho University, Can Tho, Vietnam}

\date{}

\begin{document}

\maketitle

\begin{abstract}
\nd Standard automated sewer pipe severity assessment relies on direct image classification, creating a "black box" where the link between visual defects and final severity scores remains implicit. This study introduces a modular, fuzzy rule-based neuro-symbolic framework that bridges this gap by decoupling neural perception from symbolic reasoning. The perception module utilizes a Swin Transformer to predict 14 multilabel inspection CODE degrees directly from images. For reasoning, a DT, specifically Weka's J48, algorithm is trained on ground-truth CODEs and severity labels, and its paths are converted into 19 fixed IF--THEN rules. Inference operates via fuzzy logic: t-norm activations from CODE conditions are weighted by rule confidence and combined with corresponding s-norms to produce interpretable class evidence. We assessed Product, {\L}ukasiewicz, and Hamacher operator pairs using a dataset of 3,244 images spanning five highly imbalanced severity classes. Ground-truth labels were robustly generated via consensus from five independent large language models analyzing original inspector notes. Our results show an improvement of accuracy, balanced accuracy, Macro F1 and MCC by 17.9\%, 12.2\%, 23.0\%, and 17.3\%, respectively, over image-only based classification.
%
% The Product and Hamacher soft-degree variants achieve an accuracy of 0.7485, a balanced accuracy of 0.5272, a macro F1 of 0.5352, and an MCC of 0.4901. Relative to the image-only baseline, these values correspond to improvements of 17.9\%, 12.2\%, 23.0\%, and 17.3\%, respectively.
%
% Furthermore, compared to image–CODE fusion, our framework yielded superior balanced accuracy and equivalent macro F1, despite fusion maintaining higher overall accuracy. 
Overall, the framework combines competitive class-balanced performance with traceable reasoning from predicted CODE degrees to rule supports and severity evidence.
\end{abstract}


% \begin{highlights}
% \item A modular neuro-symbolic framework for predicting sewer pipe severity prediction from images.
% \item Multilabel inspection CODE degrees prediction that bridges visual perception and fuzzy rule reasoning. 
% \item A fixed, automatically constructed rule base enabeling transparent severity classification.
% \item Balanced accuracy and Macro F1 improve by 16.2\% and 23.4\% over image-only severity classification 
% \end{highlights}

%\item Proposes a modular neuro-symbolic framework for image-based sewer pipe severity prediction.
%\item Predicts multilabel inspection CODE degrees bridging visual perception and fuzzy rule reasoning.
%\item Ensures transparent severity classification through an automatically constructed, fixed rule base.
%\item Outperforms image-only classification, improving accuracy, balanced accuracy, Macro F1 and MCC by 17.9\%, 12.2\%, 23.0\%, and 17.3\%, respectively.
%\nocite{*}

% Keywords

\maketitle

\section{Introduction}
\label{sec1}

\nd Sewer networks are essential urban infrastructure assets that support public health, environmental protection, and urban service continuity. Their deterioration may lead to hydraulic inefficiency, structural failure, groundwater infiltration, blockage, and costly emergency interventions among others~\citep{Fenner00, Tscheikner19, Ana10}. Therefore, regular inspection and reliable condition assessment are important for maintenance planning, risk reduction, and rehabilitation prioritization. In sewer inspection practice, visual observations are commonly translated manually into defect descriptions, standardized defect CODEs, and severity levels. Among these outputs, the severity prediction is particularly important because it provides a compact decision variable for prioritizing maintenance actions and identifying pipe segments that require further attention.

Recent advances in computer vision have encouraged the development of automated sewer pipe inspection methods based on deep neural networks. Image-based models can learn visual patterns associated with pipe defects and can be trained to predict either defect categories or severity classes~\citep{Haurum20, Kumar18, Wang21}. However, pipe severity prediction from sewer inspection images still remains a challenging task.  CCTV images are often affected by low illumination, blur, water, deposits, occlusion, viewpoint changes, and visual noise. Moreover, sewer defects are visually diverse and may co-occur in the same image. This means that visual evidence is often multilabel at the defect level, whereas the final output is usually represented as a single severity class. The task is further complicated by high class imbalance because low-severity cases are much more frequent than critical cases in many inspection datasets. Let us note that in practical sewer inspection, severity assessment is rarely performed directly from raw images. Instead, inspectors first identify standardized defect CODEs that describe observable structural conditions and subsequently determine the overall severity according to inspection guidelines. Defect CODEs therefore provide a natural semantic representation that bridges low-level visual observations and high-level maintenance decisions. This practical workflow suggests that explicit modelling of intermediate defect semantics may offer a more transparent alternative to direct image-to-severity prediction.

% A key difficulty is that defect recognition and severity prediction are related but distinct tasks.

Although recent deep learning models have achieved encouraging predictive performance, most approaches formulate pipe severity prediction as a direct image classification problem. However, defect recognition and severity prediction are related but fundamentally different tasks. A visual model may identify local defect evidence, but the assignment of a severity level often depends on how several defect indicators are combined. For instance, the presence, absence, and interaction of different defect CODEs may lead to different severity decisions. Therefore, pipe severity prediction is not only a visual classification problem, but also a reasoning problem over intermediate defect semantics. Direct image-to-severity models may provide useful predictive effectiveness, but usually do not explicitly represent the defect-level reasoning process that links visual observation to severity outcomes. Consequently, the reasoning process that links visual evidence to severity decisions remains largely implicit, making it difficult to explain predictions or verify which defect evidence supports a particular maintenance recommendation.

To address this issue, this paper proposes a fuzzy rule-based neuro-symbolic approach for pipe severity prediction in sewer networks. The central idea is to introduce defect CODEs as an intermediate semantic representation between raw sewer pipe images and final severity classes. Instead of directly mapping an image to a severity label, the proposed framework first predicts multilabel defect CODE degrees from the inspection image and then infers the severity degree through a fixed symbolic rule base. This design separates neural visual perception from symbolic severity reasoning, while allowing the final decision to be traced through predicted CODEs, activated rules, rule confidence values, and class-level evidence scores. That is, every severity prediction can be traced through predicted defect CODEs, activated rules, rule confidence values, and aggregated class evidence, providing transparent decision support for sewer inspection.

The neural component is implemented as an image-based multilabel CODE predictor using a Swin Transformer backbone~\cite{liu21}, reflecting the fact that multiple sewer defects may co-occur in the same inspection image. 
The symbolic component is obtained from a single Decision Tree (DT)~\cite{Mitchell97,Quinlan87} trained on ground-truth CODE annotations and final severity labels. Specifically, we used Weka's J48 implementation of C4.5~\cite{Quinlan93,Witten11}). The extracted root-to-leaf paths form IF--THEN rules that encode explicit relationships between defect conditions and severity classes. During inference, the rule base remains fixed and is applied to the CODE degrees produced by the neural model to infer the final severity degree. 

% As neural predictions are inherently imprecise, we rely on fuzzy rule activations that permit graded rule satisfaction and enables graded CODE evidence to be propagated through the symbolic reasoning layer. The detailed formulation of the proposed framework, including fuzzy rule activation and class-level aggregation, is presented in Section~\ref{sec3}.

It is worth noting that the proposed approach should be carefully distinguished from end-to-end neuro-symbolic learning systems. In this study,  the symbolic rule base is not jointly optimized with the neural model, and no rule-loss backpropagation is used. Instead, the term neuro-symbolic refers to the integration of neural multilabel CODE prediction with a fixed symbolic rule-based reasoning layer. The neural component performs image-based perception, whereas the symbolic component performs interpretable severity inference. This separation is intentional because the main objective is to improve transparency in pipe severity prediction by making the intermediate defect representation and rule-base reasoning process explicit.

In summary, the contributions of this study are fivefold. 
\begin{enumerate}
    
\item We propose a fuzzy rule-based neuro-symbolic framework for pipe severity prediction in sewer networks, in which defect CODEs are used as an intermediate semantic representation between sewer pipe images and final severity classes. 

\item We construct severity labels from inspector-written observation texts using a consensus protocol based on five independent large language models, where CODE annotations are not used during severity labelling. 

\item We formulate image-based defect CODE recognition as a multilabel classification task, reflecting the practical situation in which multiple pipe defects may co-occur in a single inspection image. 

\item We extract a symbolic rule base from ground-truth CODE annotations and severity labels using a single DT, viz.~Weka's J48 algorithm, and use the resulting IF--THEN rules for fuzzy rule-based severity reasoning with t-norm activation, rule confidence weighting, and s-norm aggregation~\citep{Klir95,Straccia08a,Straccia13}.

\item Finally, we evaluate the proposed framework against the image-only baseline and further analyse its behaviour through Oracle CODE + Rule reasoning, threshold sensitivity analysis, and qualitative explanation cases.

\end{enumerate}

\nd Beyond predictive performance, the proposed framework provides an interpretable reasoning process that can assist engineers in understanding, validating, and refining maintenance decisions. The modular architecture also facilitates future adaptation to different visual backbones or inspection standards without redesigning the complete framework.

In the following, we proceed as follows. In Section~\ref{sec2}, we review related literature. In Section~\ref{sec3} we illustrate our framework, while in Section~\ref{sec4} we discuss our experimental setup. In Section~\ref{sec5} we discuss and present our results. Finally, Section~\ref{sec6} concludes the paper.

\section{Related Work}
\label{sec2}

%\subsection{Vision-based sewer inspection and condition assessment}

\nd Automated interpretation of sewer imagery has evolved from feature-engineered pattern recognition to deep visual learning. Early studies combined image preprocessing and engineered descriptors with neural and neuro-fuzzy classifiers to distinguish cracks, holes, joints, and other pipe defects ~\citep{Chae01, Sinha06, Sinha02}. These methods demonstrated the relevance of fuzzy fusion and classification to noisy pipe imagery, but their decisions were directed primarily towards defect recognition from predefined visual features. With the development of deep learning, convolutional models increasingly replaced handcrafted pipelines. Kumar et al.~\citep{Kumar18} used an ensemble of binary convolutional neural networks to recognize multiple defect types in CCTV images, while Li et al.~\citep{Li19} introduced hierarchical classification and resampling to address severe class imbalance. Hassan et al.~\citep{Hassan19} further developed a convolutional framework for sewer defect classification and condition assessment. These studies established the effectiveness of learned visual representations, but the connection between recognized defect evidence and a final condition decision generally remained embedded in the predictive model.

The release of Sewer-ML shifted attention towards large-scale multilabel sewer defect recognition~\citep{Haurum20}. Its formulation is particularly relevant because several inspection CODEs may co-occur in the same image. Subsequent work has modelled dependencies among defects and pipe properties. The Cross-Task Graph Neural Network jointly predicts defects, water level, pipe material, and pipe shape by exchanging information across task-specific outputs~\citep{Haurum22}.  These approaches exploit label or task relationships within an end-to-end neural predictor. However, their semantic outputs are not subsequently evaluated by an explicit rule base that exposes how particular inspection CODE combinations support a severity class.

Research on condition assessment has also considered defect localization and geometric severity quantification. Wang et al.~\citep{Wang21} combined defect detection, segmentation, geometric measurements, and standard-based criteria to estimate defect severity and overall sewer condition. Zhou et al.~\citep{Zhou22} used pixel-level semantic segmentation to derive defect type, location, geometry, and severity. Such methods are valuable when severity can be determined from measurable defect extent. Nevertheless, they differ from the problem considered here, where a single severity class is inferred from the joint presence and absence of several standardized inspection CODEs. Comparatively less attention has been given to making the CODE-level reasoning that connects image evidence to the final severity decision explicit.

The use of inspection CODEs as an intermediate representation is related to concept-based and neuro-symbolic learning. Concept Bottleneck Models decompose prediction into input-to-concept and concept-to-target mappings, thereby exposing human-understandable intermediate variables~\citep{Koh20}. Standardized sewer CODEs provide such semantics, but a concept bottleneck alone does not guarantee an explicit derivation because its downstream mapping may remain statistical. More broadly, neuro-symbolic systems connect neural predictions with probabilistic logic, answer set programs, differentiable first-order semantics, or learned relational reasoning, as exemplified by NeurASP, Logic Tensor Networks, and Neural Logic Machines (see, e.g.~\citep{Badreddine22,Dong19,Sarker21, Yang23}. These systems commonly support tight coupling through differentiable inference, joint learning, or symbolic supervision. In contrast, the present framework is deliberately modular: the neural component predicts multilabel inspection CODE degrees, whereas a rule base extracted separately from ground-truth CODEs and severity labels remain fixed during inference and does not impose a rule-based training loss. The term neuro-symbolic therefore denotes the composition of neural perception of basic concepts (in our case CODEs) and symbolic reasoning using these concepts (CODEs) rather than end-to-end differentiable logic learning. 

% Let us note works such as Fuzzy OWL-Boost and PN-OWL are inspired by a similar idea, i.e., has shown how graded symbolic rules can be induced and aggregated over structured ontology instances~\citep{Cardillo24,Cardillo22}.

Let us note that fuzzy reasoning is motivated because visual evidence for an inspection CODE is in fact graded: an image is about a CODE to some extent~\cite{Meghini01} rather than strictly Boolean and  t-norms and s-norms provide principled operators for combining antecedents and aggregating alternative rule supports under many-valued semantics~\citep{Klir95}. Work such as Fuzzy OWL-Boost and PN-OWL has shown how graded symbolic rules can be induced and aggregated over structured ontology-based data~\citep{Cardillo24,Cardillo22}.  Fuzzy methods also have precedents in sewer inspection: Chae and Abraham~\citep{Chae01} fused neural defect recognizers using fuzzy logic; later neuro-fuzzy classifiers combined membership functions with engineered visual features~\citep{Sinha06,Sinha02}. These studies establish the relevance of fuzzy modelling, but do not combine modern multilabel CODE prediction with a fixed and explicitly traceable severity rule base as we do. 

% Here, crisp DT paths define the rules, while predicted CODE degrees are combined by a t-norm, weighted by empirical rule confidence, and aggregated into class evidence by the corresponding s-norm. Fuzziness thus preserves graded evidence at the neuro-symbolic interface rather than learning the visual classifier or rule structure.

Taken together, prior work leaves three research streams only weakly connected: multilabel recognition of sewer inspection CODEs, interpretable mapping from CODE combinations to severity, and fuzzy propagation of neural CODE degrees prediction through symbolic rules. To the best of our knowledge, these elements have not been jointly investigated for image-only sewer severity prediction.

% with a numerical trace from predicted CODE degrees to rule activations, confidence-weighted supports, class evidence, and the final decision. This study addresses this gap through a compact DT-derived rule base and fuzzy evidence aggregation. Its contribution lies not in a new visual backbone or fuzzy operator, but in an explicit and empirically evaluated semantic reasoning architecture for sewer pipe severity assessment.

\section{Proposed Methodology}
\label{sec3}

\nd This section presents the proposed fuzzy rule-based neuro-symbolic framework for pipe severity prediction in sewer networks. The objective is to define an image-based inference pipeline that first maps a sewer pipe inspection image to multilabel defect CODE degrees and then maps these degrees to a severity class through symbolic fuzzy rules. As already mentioned, unlike direct image-to-severity classification, the proposed formulation makes the intermediate defect representation and the subsequent rule-based reasoning process explicit.

Specifically, given a sewer inspection image $x_i$, the objective is to predict its severity class $y_i\in\mathcal{Y}$. Rather than directly learning the mapping $x_i\rightarrow y_i$, the proposed framework first estimates a multilabel defect CODE degree vector
\(\vec{\mu}_i=(\mu_{i1},\ldots,\mu_{im})\in[0,1]^m\), where each $\mu_{ij}$ is the predicted degree of CODE $j$ for image $x_i$. The severity prediction is then obtained by applying fuzzy rule reasoning to \(\vec{\mu}_i\).
Accordingly, the framework contains two separate components. The neural component performs multilabel visual perception by estimating defect CODE degrees from the input image, whereas the symbolic component performs severity reasoning by applying a rule base extracted from ground-truth CODE annotations and final severity labels. The rule base is fixed after extraction and is not optimized jointly with the neural model.

\subsection{Framework overview}
\label{subsec3.1}

\nd The proposed framework consists of an offline symbolic rule base construction phase and an online image-based inference phase. In the offline phase, a DT classifier is trained using ground-truth defect CODE annotations and final severity labels. Each root-to-leaf path of the trained tree is then converted into an IF--THEN rule. The extracted rules define interpretable mappings from CODE-based defect conditions to severity classes. Each rule is also associated with a confidence value computed from the empirical class distribution of its corresponding leaf node.

\begin{figure*}%[]
  \centering
\includegraphics[width=\linewidth]{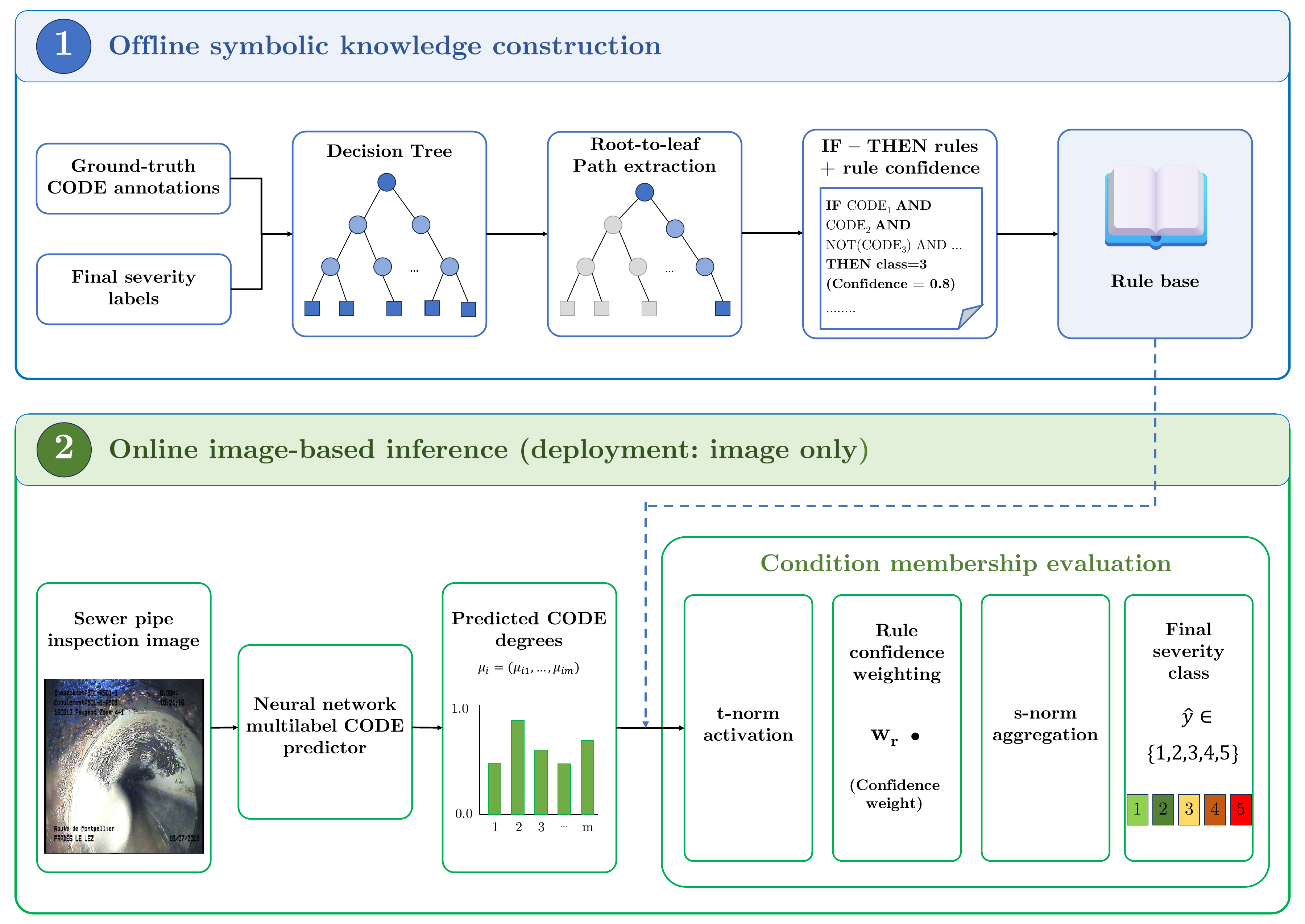}
    \caption{Overall framework of the proposed fuzzy rule-based neuro-symbolic approach for sewer pipe severity prediction. In the offline phase, ground-truth CODE annotations and severity labels are used to train a DT, from which root-to-leaf paths are extracted as IF--THEN rules with associated confidence values. The resulting rule base is fixed after extraction and used for the final severity class prediction. Separating rule extraction from inference preserves the interpretability of the symbolic knowledge while avoiding repeated rule learning during model deployment.} \label{fig:overall_framework}
\end{figure*}

In the online phase, the input is a sewer pipe inspection image. A Swin Transformer-based multilabel model first predicts a degree vector over the defect CODEs used by the final rule base. These CODE degrees are then fed into the symbolic rule base. For each rule, CODE conditions are transformed into fuzzy membership values, combined using a so-called t-norm, and weighted by the rule confidence. Finally, the confidence-weighted rule degrees are aggregated for each severity class using a s-norm (see, e.g.,~\cite{Klir95} for an introduction to Fuzzy Logic).\footnote{We refer the reader to Appendix~\ref{app1} for more details about the extracted rule base.} The predicted severity class is selected as the class with the highest aggregated degree.
Figure~\ref{fig:overall_framework} summarizes the proposed framework, which separates offline symbolic rule base construction from online image-based severity inference.

This design follows the central hypothesis of the paper: defect CODEs provide a meaningful semantic layer through which the relationship between visual defects and severity decisions can be made explicit.

\subsection{Problem formulation } \label{subsec3.2}
\nd Let us now formalise the data representation and prediction task. Specifically, let
\[
\mathcal{D}=\{(x_i,o_i,\vec{z}_i,y_i)\}_{i=1}^{N}
\]
\nd denote the dataset, where $x_i$ is a sewer pipe inspection image, $o_i$ is the corresponding inspector-written observation, $\vec{z}_i\in\{0,1\}^{m}$ is the ground-truth multilabel defect CODE vector, and $y_i\in\mathcal{Y}$ is the final severity class. The severity label space is defined as
\[
\mathcal{Y}=\{1,2,3,4,5\}\ ,
\]
\nd where larger values indicate more severe pipe conditions (see also Table~\ref{tab:inspection_example} for a dataset element example).

The observation \(o_i\) is used exclusively during offline dataset construction to derive \(y_i\) through the multi-LLM consensus protocol described in Section~\ref{subsec4.1} later on. The ground-truth CODE vector
\(\vec{z}_i\) is used for multilabel supervision and Decision
Tree rule extraction. During deployment, only the image \(x_i\) is
available. The objective is to realize an image-based severity predictor
\[
f:x_i \mapsto \hat{y}_i \ ,
\]
\nd by decomposing this mapping into two interpretable stages:
\[
x_i \rightarrow \vec{\mu}_i \rightarrow \hat{y}_i \, 
\]
\nd with 
\[
\vec{\mu}_i=(\mu_{i1},\mu_{i2},\ldots,\mu_{im})\in[0,1]^m \ ,
\]

\nd where $\vec{\mu}_i$ is the predicted CODE degree vector, and $m$ denotes the number of CODE labels retained as the semantic interface between the neural perception module and the fuzzy rule layer.
Specifically, $f=g\circ h$ with
\[
h:x_i\rightarrow\vec{\mu}_i \ , and
\]
\[
g:\vec{\mu}_i\rightarrow\hat y_i \ .
\]
\nd This decomposition explicitly separates perception from reasoning, allowing intermediate defect semantics to be inspected before the final severity decision is produced.
The first stage corresponds to neural visual perception, whereas the second stage corresponds to fuzzy symbolic reasoning for predicting the severity degree. 

\subsection{Multilabel defect CODE prediction}
\label{subsec3.3}

\nd Defect CODEs provide a standardized semantic representation of pipe conditions. Since multiple defects may occur in a single inspection image, CODE prediction is formulated as a multilabel classification problem. 
This formulation is more appropriate than multiclass classification because the presence of one defect CODE does not exclude the presence of another.

The original dataset contains a larger set of defect CODE labels. After rule extraction, only the CODE variables appearing in the selected symbolic rules are retained as the semantic interface (a.k.a. \textit{semantic bottleneck}) between the neural perception module and the symbolic reasoning module. The concrete CODE set used in the experiments is reported in Section~\ref{subsec4.1} and Appendix~\ref{app1}. The multilabel model outputs the CODE degree vector $\vec{\mu}_i$ defined in Section~\ref{subsec3.2}. Each element $\mu_{ij}$ is obtained from the sigmoid output of the multilabel head and denotes the degree of truth to which image $x_i$ visually \emph{is about} or \emph{satisfies} CODE $j$ (as such, we follow conceptually the multimedia model described in~\cite{Meghini01}). These values are not final severity predictions; instead, they provide semantic evidence for the fuzzy rule layer. They are used then as fuzzy degrees in the reasoning process. Consequently, the neural network is trained to predict defect semantics rather than severity labels directly.
%rather than as calibrated probabilities.

The model is trained using Binary Cross-Entropy loss:
\[
\mathcal{L} = -\frac{1}{N_{tr}} \sum_{i=1}^{N_{tr}} \sum_{j=1}^{m} \left[ w_j z_{ij}\log\mu_{ij} + (1-z_{ij})\log(1-\mu_{ij}) \right],
\]
where $z_{ij}\in\{0,1\}$ is the ground-truth label of CODE $j$ for image $x_i$, and $w_j$ is the positive class weight used to mitigate CODE imbalance. This weighting reduces the influence of severe label imbalance among the defect CODEs.

% \todo[inline]{UMBERTO: Thai, insert table about CODEs distribution showing unbalance}
\begin{table}[t]
\caption{Distribution of the 14 CODEs retained by the rule base. Since the task is multilabel, the CODE counts are not mutually exclusive.}
\label{tab:code_distribution}
\centering
\scriptsize
\setlength{\tabcolsep}{2pt}
\begin{tabular*}{\textwidth}{@{\extracolsep{\fill}}lrrrrr@{}}
\toprule
CODE & Train & Val. & Test & Total & Prevalence (\%) \\
\midrule
\texttt{BCE} & 512 & 66 & 65 & 643 & 19.82 \\
\texttt{BDD} & 221 & 29 & 28 & 278 & 8.57 \\
\texttt{BAA} & 202 & 26 & 25 & 253 & 7.80 \\
\texttt{BDC} & 70  & 9  & 9  & 88  & 2.71 \\
\texttt{BAB} & 64  & 8  & 8  & 80  & 2.47 \\
\texttt{BAF} & 60  & 8  & 7  & 75  & 2.31 \\
\texttt{BAC} & 59  & 8  & 7  & 74  & 2.28 \\
\texttt{BBA} & 47  & 6  & 6  & 59  & 1.82 \\
\texttt{BAJ} & 26  & 3  & 3  & 32  & 0.99 \\
\texttt{DBA} & 22  & 3  & 3  & 28  & 0.86 \\
\texttt{BBB} & 22  & 3  & 2  & 27  & 0.83 \\
\texttt{BAI} & 21  & 2  & 3  & 26  & 0.80 \\
\texttt{BAG} & 17  & 2  & 2  & 21  & 0.65 \\
\texttt{BBC} & 16  & 2  & 2  & 20  & 0.62 \\
\bottomrule
\end{tabular*}
\end{table}
As shown in Table~\ref{tab:code_distribution}, the retained CODEs are highly imbalanced, with overall prevalence ranging from 0.62\% for \texttt{BBC} to 19.82\% for \texttt{BCE}.

\nd Only the neural multilabel CODE prediction model is optimized by gradient descent. As already mentioned, the symbolic rule base is extracted once from ground-truth CODE annotations and remains fixed throughout training and inference. This multilabel prediction module therefore serves as the perception component of the proposed neuro-symbolic framework, providing the semantic evidence required for the subsequent fuzzy reasoning process, which is described next.

\subsection{Symbolic rule extraction}
\label{subsec3.4}

\nd The symbolic knowledge base is extracted from a single DT classifier trained on ground-truth CODE annotations and final severity labels. Spcifically, we used  Weka's J48 implementation of C4.5. The classifier is pruned using the scheme
\begin{center}
\texttt{weka.classifiers.trees.J48 -C 0.25 -M 2},
\end{center}
\nd where the confidence factor for pruning is 0.25 and the minimum number of training instances per leaf is 2. 

% Now, let
% \[
% \vec{u}_i\in\{0,1\}^{n}
% \]
% denote the candidate CODE feature vector used for rule extraction.
% % In contrast to the reduced CODE representation used by the neural predictor, $\vec{u}_i$ contains all candidate CODE variables available for symbolic rule extraction.
% Here, $n$ denotes the number of candidate CODE features available for rule extraction. J48 learns a mapping
% \[
% g_{\mathrm{J48}}:\vec{u}_i\mapsto y_i.
% \]
% Each path terminating at a J48 leaf is converted into one symbolic rule. A rule $R_r$ is written as
% \[
% R_r: \quad \text{IF } \bigwedge_{l=1}^{L_r} \gamma_{rl} \text{ THEN } y=c_r,
% \]
% where $\gamma_{rl}$ is the $l$-th CODE-based condition, $L_r$ is the number of antecedent conditions, and $c_r\in\mathcal{Y}$ is the severity class predicted by the corresponding leaf. 

Now, let
\[
\vec{u}_i\in\{0,1\}^{n}
\]
denote the candidate CODE feature vector used for rule extraction. 
% In contrast to the reduced CODE representation used by the neural predictor, $\vec{u}_i$ contains all candidate CODE variables available for symbolic rule extraction.
Here, $n$ denotes the number of candidate CODE features available for rule extraction. The algorithm learns a mapping
\[
g_{\mathrm{DT}}:\vec{u}_i\mapsto y_i.
\]
Each root-to-leaf path of the trained tree is converted into one symbolic rule. A rule $R_r$ is written as
\[
R_r: \quad \text{IF } \bigwedge_{l=1}^{L_r} \gamma_{rl} \text{ THEN } y=c_r,
\]
where $\gamma_{rl}$ is the $l$-th CODE-based condition, $L_r$ is the number of antecedent conditions, and $c_r\in\mathcal{Y}$ is the severity class predicted by the corresponding leaf.

Let $N_r$ denote the number of training samples reaching that leaf and $E_r$ the number not belonging to its predicted class, following Weka's leaf notation $(N_r/E_r)$. The confidence of rule $R_r$ is defined as
\[
\operatorname{Conf}_r = \frac{N_r-E_r}{N_r}.
\]
The complete rule base is denoted by
\[
\mathcal{R}=\{R_1,R_2,\ldots,R_K\}.
\]
Let $X$ be a variable that denotes an inspection image $x_i \in \mathcal{X}$, let $b_{rl}(X)$ denote the $l$-th CODE-based body literal of rule $R_r$, and let $\operatorname{sev}_{c_r}(X)$ denote the severity predicate associated with class $c_r$. A symbolic rule can also be represented as (written in Datalog style~\cite{Ullman89}):
\[
R_r: \operatorname{sev}_{c_r}(X) \leftarrow b_{r1}(X),b_{r2}(X),\ldots,b_{rL_r}(X),
\]
where $L_r$ is the number of literals in rule $R_r$, the right-hand side is the rule body and the left-hand side is the rule head. The comma on the right-hand side is interpreted as conjunction. Therefore, the rule states that an image $x_i$ denoted by $X$ is assigned to severity class $c_r$ when all CODE-based body literals are satisfied. For example, an extracted rule can be written as
\[
R_r: \operatorname{sev}_{2}(X) \leftarrow \operatorname{bac}_{\mathrm{absent}}(X),
\operatorname{bdd}_{\mathrm{present}}(X).
\]

\nd This rule indicates that a specific pattern of absent and present CODE conditions provides symbolic evidence for severity class 2. The confidence assigned to each rule is computed from the class distribution of the training instances reaching the corresponding leaf of the DT.

The extracted rule base remains fixed during inference. No ensemble or additional voting layer is used for symbolic knowledge extraction. This choice keeps the symbolic knowledge base compact and avoids an additional ensemble voting layer, thereby preserving a direct rule-level inference trace.
%\footnote{Other rule extraction methods may be applied. We leave the comparison with other rule extraction methods, such as RIPPER~\cite{Cohen95} for future work.} 
%
After rule extraction, the CODEs appearing in $\mathcal{R}$ define the final semantic interface used by the multilabel prediction model. 

Overall, the extracted rule base $\mathcal{R}$ from our experimental dataset contains 19 symbolic rules involving 14 retained defect CODE conditions (see Appendix~\ref{app1}).

Of course, other rule extraction methods may be applied, but we leave the comparison with other rule extraction methods, such as RIPPER~\cite{Cohen95} for future work.

\subsection{Fuzzy rule-based severity reasoning}
\label{subsec3.5}

\nd At inference time, given an image $x_i$, the symbolic rules are evaluated using the predicted CODE degree vector
\[
\vec{\mu}_i=(\mu_{i1},\mu_{i2},\ldots,\mu_{im}),
\]
where $m=14$ corresponds to the number of retained defect CODEs after symbolic rule extraction. Each rule condition is mapped to a condition degree in $[0,1]$, representing the degree to which the condition is satisfied by the neural CODE predictions. For a positive body atom of the form $b_{rl}(x_i) = \text{CODE}_{j\mathrm{present}}$, the corresponding condition degree is
\[
d_{irl} \coloneq \mu_{ij}.
\]
For a `negative' body atom of the form $b_{rl}(x_i) = \text{CODE}_{j\mathrm{absent}}$, the corresponding condition degree is
\[
d_{irl} \coloneq 1-\mu_{ij}.
\]

\nd To analyse the effect of fuzzy connective t-norm/s-norm selection, the rule reasoning layer is instantiated with three well-known operator settings: Product, {\L}ukasiewicz, and Hamacher~\cite{Klir87}. Different fuzzy operators capture different semantics for conjunction and disjunction under fuzzyness. Evaluating multiple operator settings allows us to assess the robustness of the proposed reasoning mechanism with respect to the choice of fuzzy logic operators. Specifically, let $\omega\in\{\mathrm{P}, \mathrm{L}, \mathrm{H}\}$ denote the selected operator setting. For each setting, a t-norm $t_\omega$ is used to compute rule activation, while a s-norm $s_\omega$ is used to aggregate confidence-weighted rule supports at the class level.

For \(a,b\in[0,1]\), the \emph{Product} setting is defined as
\[
t_{\mathrm{P}}(a,b)=ab,
\qquad
s_{\mathrm{P}}(a,b)=a+b-ab.
\]

\nd The \emph{{\L}ukasiewicz} setting is defined as
\[
t_{\mathrm{L}}(a,b)=\max(0,a+b-1),
\qquad
s_{\mathrm{L}}(a,b)=\min(1,a+b).
\]

\nd The \emph{Hamacher} setting is defined as
\[
t_{\mathrm{H}}(a,b)=
\frac{ab}{a+b-ab},
\qquad
s_{\mathrm{H}}(a,b)=
\frac{a+b-2ab}{1-ab},
\]
with \(t_{\mathrm{H}}(0,0)=0\) and \(s_{\mathrm{H}}(1,1)=1\).

Given a rule \(R_r\) with condition degrees
\[
d_{ir1},d_{ir2},\ldots,d_{irL_r},
\]
the activation degree is computed as
\[
A_{ir}^{(\omega)}
=
t_\omega(d_{ir1},d_{ir2},\ldots,d_{irL_r}),
\]
where \(t_\omega\) denotes the selected fuzzy conjunction operator. 
%For Product and {\L}ukasiewicz, the \(n\)-ary forms are directly used. For Hamacher, the \(n\)-ary form is computed recursively.

The confidence-weighted support for rule $r$ is:
\[
\operatorname{Supp}_{ir}^{(\omega)}=A_{ir}^{(\omega)}\cdot\operatorname{Conf}_r.
\]
This weighting explicitly incorporates both neural evidence and symbolic reliability into the reasoning process.
This formulation prevents weakly activated rules or low-confidence rules from dominating the final decision. 

Moreover, for each severity class $c$, let
\[
\mathcal{R}_c=\{R_r\in\mathcal{R}:c_r=c\}
\]
denote the set of rules whose consequent is class $c$. The aggregated severity degree for class $c$ is computed using the selected fuzzy disjunction operator $s_\omega$:
\[
E_i^{(\omega)}(c) = s_\omega \left( \{\operatorname{Supp}_{ir}^{(\omega)} \mid R_r \in \mathcal{R}_c\} \right).
\]
The resulting class evidence represents the accumulated support provided by all rules whose consequent corresponds to severity class $c$.
If no rule supports class $c$, then $E_i^{(\omega)}(c)=0$. The final severity prediction under operator setting $\omega$ is obtained as
\[
% \hat{y}_i^{(\omega)}=\arg\max_{c\in\mathcal{Y}} s_i^{(\omega)}(c).
\hat{y}_i^{(\omega)}=\arg\max_{c\in\mathcal{Y}} E_i^{(\omega)}(c).
\]
For explanatory purposes, the aggregated severity degrees can be normalized as
\[
\widetilde{E}_i^{(\omega)}(c) = \frac{E_i^{(\omega)}(c)}{\sum_{c' \in \mathcal{Y}} E_i^{(\omega)}(c') + \epsilon},
\]
where $\epsilon=10^{-6}$ avoids division by zero. These normalized values are not used to modify the prediction; they only provide a relative view of the evidence assigned to each severity class. 

The reasoning process can then be summarized as follows:
\[
\vec{\mu}_i \rightarrow \{d_{irl}\} \rightarrow \{A_{ir}^{(\omega)}\} \rightarrow \{\operatorname{Supp}_{ir}^{(\omega)}\} \rightarrow \{E_i^{(\omega)}(c)\}_{c \in \mathcal{Y}} \rightarrow \hat{y}_i^{(\omega)}.
\]
\nd For a crisp variant of the rule interface, CODE degrees may be thresholded before rule evaluation ($\tau \in (0,1]$):
\begin{equation} \label{tau}
\hat{z}_{ij}^{(\tau)} = \begin{cases} 1 & \mu_{ij}\geq \tau \\ 0 & \mu_{ij}<\tau \ , \end{cases}
\end{equation}
\nd where $\mu_{ij}$ denotes the predicted degree of CODE $j$ for image $i$. 
The threshold values used in the experiments are specified in Section~\ref{subsec4.5}. The same symbolic reasoning procedure is then applied by replacing $\mu_{ij}$ with $\hat{z}_{ij}^{(\tau)}$ in the condition degree computation. These two modes allow the experiments to compare fuzzy degree-based reasoning versus crisp threshold-based reasoning. %is more compatible with the extracted symbolic rules.
The resulting inference trace forms the basis of the explanation mechanism described in Section~\ref{subsec3.6}.

\subsection{Explanation mechanism and inference trace}
\label{subsec3.6}

\nd The proposed framework provides an intrinsic explanation of every severity prediction by exposing the complete semantic reasoning process from defect CODE recognition to symbolic rule inference. The explanation is generated as part of the inference process itself rather than by an external post-hoc explanation method. For an input image, the model produces an explanation trace consisting of the predicted CODE degrees, the activated IF--THEN rules, rule activation degrees, rule confidence values, confidence-weighted supports, and class-wise severity degrees. The local explanation chain is as follows:
\begin{quote}
    Image $\rightarrow$ Predicted CODE degrees $\rightarrow$ Activated IF--THEN rules $\rightarrow$ Rule supports $\rightarrow$ Class-wise severity degrees $\rightarrow$ Final severity.
\end{quote}
The predicted CODE degrees indicate which defect-related semantic concepts are visually supported by the image. The activated rules show how these CODEs satisfy symbolic defect conditions. The confidence values indicate the empirical reliability of the rules extracted from the DT, while the class-wise fuzzy aggregation degrees show how evidence is accumulated before the final severity class is selected.

This explanation mechanism is aligned with the central objective of the paper: to provide an interpretable alternative to direct image-only severity classification. The framework does not merely output a severity label; it also exposes the intermediate semantic predictions and the symbolic rules that contribute to the final decision. This may support inspection scenarios in which severity predictions need to be reviewed, validated, or compared with defect-based engineering knowledge.

\section{Experimental Setup}
\label{sec4}

\nd This section describes the experimental setup used to evaluate the proposed fuzzy rule-based neuro-symbolic framework. The methodological formulation of the proposed model is given in Section~\ref{sec3}; therefore, this section focuses on the dataset, leakage control, compared methods, implementation details, inference settings, and evaluation metrics.

\subsection{Dataset and split}
\label{subsec4.1}

\nd The dataset consists of sewer pipe inspection records collected from a sewer inspection project in France.\footnote{The dataset is proprietary and unfortunately can not yet be disclosed.} Each sample (see Table~\ref{tab:inspection_example}) contains a sewer pipe inspection image, an inspector-written observation, one or more defect CODE annotations, and a final severity label. The observation text  describes the inspected pipe condition in natural language (in French), whereas the CODE annotations provide a standardized defect representation. Each inspection record therefore combines visual information, textual descriptions, standardized defect annotations, and an associated severity assessment.

% Table~\ref{tab:inspection_example} provides an illustrative inspection record showing the relationship among the sewer image, the inspector-written observation, the standardized defect CODE annotations, and the corresponding severity label.

\begin{table}[ht]
\centering
\caption{Illustrative example of a sewer pipe inspection record.}
\label{tab:inspection_example}
\small
\setlength{\tabcolsep}{2pt}
\renewcommand{\arraystretch}{1.2}
\begin{tabular}{
    >{\centering\arraybackslash}p{0.20\linewidth}
    p{0.30\linewidth}
    >{\centering\arraybackslash}p{0.25\linewidth}
    >{\centering\arraybackslash}p{0.10\linewidth}
}
\toprule
\textbf{Image} & \textbf{Observation} & \textbf{CODEs} & \textbf{Sev.} \\
\midrule
\raisebox{-.7\height}{\includegraphics[width=1\linewidth]{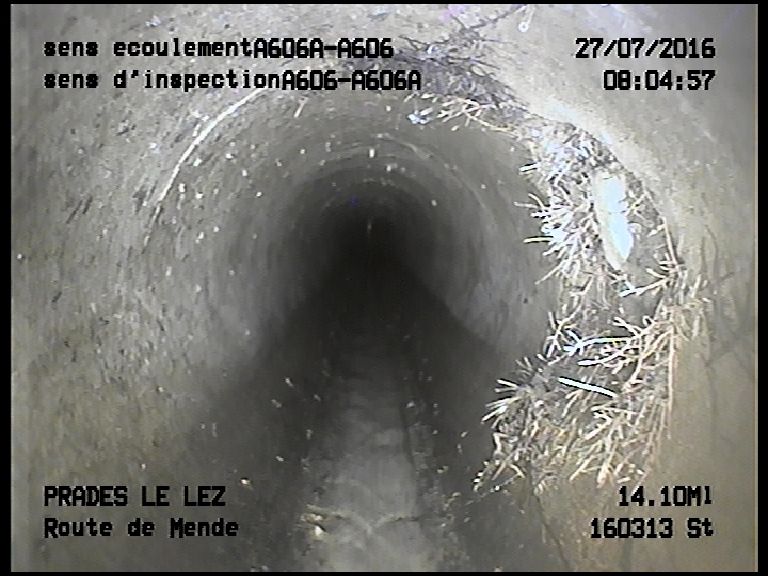}}
& Effondrement partiel à 3h Présence de radicelles & \texttt{BAC}, \texttt{BBA} & 4 \\
\bottomrule
\end{tabular}
\end{table}

\nd The final severity label belongs to five ordered classes:
\[
\mathcal{Y}=\{1,2,3,4,5\},
\]
where larger values indicate more severe pipe conditions (the severity label summarizes the overall condition on the \textit{five-level severity scale.}).

% The example illustrates that the observation text  provides a natural-language description (in French) of the inspected condition, whereas the CODE annotations encode standardized defect semantics. The severity label summarizes the overall condition on the \textit{five-level severity scale.}

Severity labels were constructed from the inspector-written observation texts using a consensus protocol based on five independent large language models. For each unique observation $o$, each model assigned a severity score in the range $\{1,2,3,4,5\}$, let
\[
s_o^{(q)}\in\{1,2,3,4,5\}, \quad q=1,\ldots,Q,
\]
denote the score assigned by the $q$-th model, where $Q=5$. Only valid scores were used for aggregation. The consensus severity score was computed as
\[
\bar{s}_o = \frac{1}{|\mathcal{V}_o|} \sum_{q\in\mathcal{V}_o} s_o^{(q)},
\]
where $\mathcal{V}_o\subseteq\{1,\ldots,Q\}$ is the set of models that produced valid scores for observation $o$. The final severity class was obtained by standard rounding and clipping:
\[
y_o= \operatorname{clip} \left( \operatorname{round}(\bar{s}_o),1,5 \right).
\]

\nd The generated severity class was then merged back into all image records associated with the same observation. Importantly, defect CODE annotations were not used during severity labelling. This separation is necessary because CODE annotations are later used for rule extraction and multilabel CODE training.

Although the dataset contains visual, textual, and symbolic information, the deployment setting considered in this work is image-only. Thus, during inference, the evaluated image-based methods receive only the sewer pipe inspection image. Observation text is used only for offline severity-label construction. Ground-truth CODE annotations are used for multilabel supervision, DT training, and Oracle diagnostic evaluation. Neither source is available to the deployed image-based framework.

The dataset contains 3,244 images and is divided into fixed training, validation, and test sets. The split statistics are reported in Table~\ref{dataset_split}.

\newcolumntype{L}{>{\raggedright\arraybackslash}p{0.25\textwidth}}
\begin{table}[t]
\caption{Dataset split used in all experiments.}\label{dataset_split}
\begin{tabular*}{\textwidth}{@{} L L L L @{}}
\toprule
Train & Validation & Test & Total \\
\midrule
2592  & 326 & 326  & 3244  \\
\bottomrule
\end{tabular*}
\end{table}

Table~\ref{severity_distribution} reports the severity class distribution. The dataset is strongly imbalanced. Severity class 1 is dominant, accounting for most inspection images, whereas severity class 5 is rare. This imbalance is important when interpreting the results because high overall accuracy may be obtained even when minority severity classes are poorly recognized.

% Define 'L' as a left-aligned column with a specific width (e.g., 20% of text width)
\newcolumntype{L}{>{\raggedright\arraybackslash}p{0.2\textwidth}}

\begin{table}[t]
\caption{Severity class distribution across the fixed data split.}\label{severity_distribution}
\begin{tabular*}{\textwidth}{@{} L L L L L @{}}
\toprule
Severity class & Train & Validation & Test & Total \\
\midrule
1              & 1689  & 211        & 209  & 2109  \\
2              & 411   & 51         & 53   & 515   \\
3              & 354   & 42         & 47   & 443   \\
4              & 95    & 12         & 13   & 120   \\
5              & 43    & 10         & 4    & 57    \\
\bottomrule
\end{tabular*}

\end{table}

%
%
%\begin{table}[t]
%\centering % Centers the table on the page
%\caption{Severity class distribution across the fixed data split.}\label{severity_distribution}
%\begin{tabular}{@{} lcccc @{}} % l for left text, c for centered numbers
%\toprule
%Severity class & Train & Validation & Test & Total \\
%\midrule
%1              & 1689  & 211        & 209  & 2109  \\
%2              & 411   & 51         & 53   & 515   \\
%3              & 354   & 42         & 47   & 443   \\
%4              & 95    & 12         & 13   & 120   \\
%5              & 43    & 10         & 4    & 57    \\
%\bottomrule
%\end{tabular}
%\end{table}

The original dataset contains a larger set of defect CODE labels. After symbolic rule extraction, only the CODE variables appearing in the selected rule base are retained as the final image-to-rule interface. The final multilabel defect CODE set used by the neural perception module consists of 14 labels (see also Appendix~\ref{app1}):
\[
\begin{aligned} \{&\text{BCE},\text{BDD},\text{BAA},\text{BDC},\text{BAB},\text{BAF},\text{BAC},\\ &\text{BBA},\text{BAJ},\text{DBA},\text{BBB},\text{BAI}, \text{BAG}, \text{BBC}\}. \end{aligned}
\]

\nd These labels are used as intermediate semantic targets for image-based multilabel learning and as the input variables of the symbolic rule reasoning layer.

\subsection{Leakage control and evaluation protocol}
\label{subsec4.2}

\nd All experiments use the same fixed train/validation/test split. The training set is used to learn neural model parameters and to extract the symbolic rule base. The validation set is used for neural model selection and for selecting the CODE-interface configuration of each fuzzy operator. Specifically, soft-degree inference and nine thresholded configurations are compared using validation Macro F1. The selected configuration is then fixed, and the test set is used only for final evaluation. 

An image-level leakage analysis was performed before training. Two criteria were checked across the training, validation, and test sets. First, the $\texttt{image\_path}$ field was inspected to verify that no identical image path appeared in more than one split. Second, the compound identifier
\[
(\texttt{source\_file}, \texttt{photo\_id})
\]
was checked across splits. No duplicated image paths and no duplicated 

\[
(\texttt{source\_file},\texttt{photo\_id})
\]
\nd identifiers were found across the three partitions. Therefore, no image-level data leakage was detected under the adopted protocol.

% The current split does not explicitly enforce source-file-level separation. Consequently, images originating from the same source file may appear in different splits. This is a limitation of the present evaluation protocol, especially if a source file corresponds to a complete inspection video, pipe segment, or inspection session. The reported results should therefore be interpreted as image-level generalization effectiveness rather than strict source-level or inspection-session-level generalization.

\subsection{Compared methods }
\label{subsec4.3}

\nd The experiments compare three methods. The purpose is to distinguish the contribution of direct image-based severity classification, intermediate CODE prediction, and symbolic rule-based reasoning.

\paragraph{Image-only severity baseline.}
The first baseline directly maps an input sewer pipe inspection image to a severity class:
\begin{center}
   Image $\rightarrow$ Swin Transformer $\rightarrow$ Severity \ . 
\end{center}
This model represents a standard image-only multiclass severity classifier. It does not use defect CODEs or symbolic reasoning.

% \paragraph{Image + predicted CODE fusion baseline.}
% The second baseline uses predicted defect CODEs as auxiliary semantic information but does not apply symbolic rules. The visual backbone extracts image features, a multilabel branch predicts CODE information, and the predicted CODE representation is fused with the visual representation for severity classification:
% \begin{quote}
%     Image $\rightarrow$ Visual features \& predicted CODEs $\rightarrow$ Fusion classifier $\rightarrow$ Severity
% \end{quote}

% \nd This baseline evaluates whether predicted CODEs improve severity classification when used as neural feature-level information, without explicit rule-based inference.

\paragraph{Our Multilabel + Rule framework.}
Our proposed framework follows the formulation in Section~\ref{sec3}. It first predicts multilabel defect CODE degrees from the input image and then infers severity using the fixed symbolic rule base:
\begin{quote}
    Image $\rightarrow$ CODE output + fixed rules $\rightarrow$ fuzzy rule reasoning $\rightarrow$ Severity.
\end{quote}
\nd The rule layer is instantiated with three fuzzy operator settings, namely Product, {\L}ukasiewicz, and Hamacher. In each setting, the corresponding t-norm is used for rule activation, whereas the corresponding s-norm is used for class-level evidence aggregation. Rule confidence weighting is enabled in all rule-based settings. The symbolic rule base is fixed after extraction from the training set and is not updated during neural training or inference.

\paragraph{Oracle CODE + Rule.}
The Oracle CODE + Rule setting replaces image-predicted CODE degrees with ground-truth CODE annotations:
\begin{quote}
    Ground-truth binary CODE annotations  $\rightarrow$ Fixed rules $\rightarrow$ Fuzzy rule aggregation $\rightarrow$ Severity.
\end{quote}
\nd This setting is not a deployable image-only model. It estimates the upper-bound effectiveness of the symbolic reasoning module when CODE prediction is perfect and helps to identify whether effectiveness limitations mainly arise from neural CODE prediction or from the symbolic rule base.

\subsection{Implementation details}
\label{subsec4.4}

\nd All image-based models use the Swin Transformer backbone
\begin{center}
%\small
%\footnotesize
\texttt{swin\_base\_patch4\_window12\_384.ms\_in22k\_ft\_in1k}.
\end{center}

\nd Input images are resized to $384\times384$ pixels and normalized using the ImageNet channel-wise mean $(0.485,0.456,0.406)$ and standard deviation
$(0.229,0.224,0.225)$.

The multilabel CODE prediction model produces 14 logits, which are converted into CODE degrees using sigmoid activation. These sigmoid outputs are interpreted as degrees of truth for CODE presence rather than calibrated probabilities. The model is trained with binary cross-entropy with logits. To mitigate CODE-level label imbalance, positive class weights are computed from the ratio between negative and positive samples and clipped to the interval $[1,30]$.

The optimizer is AdamW with an initial learning rate of $10^{-4}$ and a weight decay of $10^{-4}$. The multilabel CODE prediction model is trained for 30 epochs with a batch size of 64. A ReduceLROnPlateau scheduler monitors validation mean average precision (mAP). The best multilabel checkpoint is selected according to validation mAP. During inference, the trained multilabel model is evaluated with a batch size of 64 and two data-loading workers. The image-only severity baseline is trained as a multiclass classifier using cross-entropy loss, and its checkpoint is selected on the validation set.

% The image-only severity baseline is trained as a multiclass classifier using cross-entropy loss. The Image + predicted CODE fusion baseline uses the same visual backbone and combines visual features with the predicted CODE representation before severity classification. This fusion model is trained with a weighted combination of severity classification loss and multilabel CODE prediction loss. For all image-based severity models, checkpoint selection is performed on the validation set.

The symbolic rule base is extracted once from a single DT trained on the training split using ground-truth CODE annotations and final severity labels. Each root-to-leaf path is converted into an IF\mbox{--}THEN rule. The predicted severity class of a rule corresponds to the leaf prediction, and the rule confidence is computed from the empirical class distribution of the corresponding leaf. As already mentioned, no Random Forest, rule-loss training, or end-to-end neuro-symbolic optimization is used.

\subsection{Rule reasoning configuration }
\label{subsec4.5}

\nd The rule reasoning mechanism follows Section~\ref{subsec3.5}. In the experiments, two inference modes are evaluated.

\paragraph{Soft-degree mode.}
In soft-degree mode, the sigmoid outputs of the multilabel CODE model are passed directly to the rule layer as predicted CODE degrees. Thus, positive and negative CODE conditions are evaluated using soft degrees derived from the neural CODE predictor.

\paragraph{Threshold mode.}
In threshold mode, predicted CODE degrees are converted into binary values before rule evaluation (see Eq.~\ref{tau}). 
%
%($\tau \in (0,1]$):
% \[
% z_{ij}^{(\tau)} = \begin{cases} 1  & \mu_{ij}\geq \tau \\ 0 \   & \mu_{ij}<\tau \ , \end{cases}
% \]
% where $\mu_{ij}$ denotes the predicted degree of CODE $j$ for image $i$. 
%
The evaluated thresholds are
\[
\tau \in \{0.1,0.2,\ldots,0.9\}.
\]
\nd This mode evaluates whether crisp CODE activation is more compatible with the extracted rules than direct soft-degree activation.

For each fuzzy operator, model selection considers ten CODE-interface candidates: one soft-degree configuration and the nine thresholded configurations above. The candidate with the highest validation Macro F1 is selected separately for Product, Hamacher, and {\L}ukasiewicz. Consequently, the selection procedure can retain direct soft-degree reasoning when it outperforms all thresholded alternatives; it does not require a threshold to be chosen. After selection, the operator-specific interface configuration is fixed before evaluation on the held-out test set.

For all rule-based experiments, rule confidence weighting is enabled. Rule activation and class-level aggregation are evaluated under the three aforementioned fuzzy operator settings: Product, {\L}ukasiewicz, and Hamacher, as defined in Section~\ref{subsec3.5}. For each setting, the corresponding t-norm computes the activation degree of each rule, while the corresponding s-norm aggregates the confidence-weighted rule supports for each severity class. 
%In the Product setting, rule activation is computed using the pure Product t-norm without length normalization. 
The final severity class is selected by the maximum raw class evidence degree rather than by the normalized relative score.

\subsection{Evaluation metrics}
\label{subsec4.6}

\nd Severity classification effectiveness is evaluated using accuracy, balanced accuracy (a.k.a.~macro recall), macro F1, %weighted F1, 
macro precision, and the multiclass Matthews Correlation Coefficient (MCC). 
Because the severity distribution is highly imbalanced, macro F1 and balanced accuracy are emphasized when comparing inference settings. MCC is reported as a complementary global measure of agreement that accounts for all cells of the confusion matrix. Macro F1 is used as the primary criterion for selecting the CODE-interface configuration on the validation set. These metrics provide a more informative assessment of minority-class recognition than accuracy alone, particularly for rare severe defects, 
as accuracy may overestimate the effectiveness of a model that mainly captures the dominant low-severity class.

The multilabel CODE prediction model is evaluated using exact match, Hamming accuracy, macro F1, and mean Average Precision (mAP). Validation mAP is used for checkpoint selection because it is more informative than exact match under multilabel imbalance.

\section{Results and Discussion}
\label{sec5}

\nd This section reports and discusses the experimental results of the proposed fuzzy rule-based neuro-symbolic framework. Model selection and final evaluation are kept separate. For each fuzzy operator, the CODE-interface configuration is selected exclusively on the validation set from soft-degree inference and nine thresholded alternatives using Macro F1. The selected configuration is then fixed before evaluation on the held-out test set. 

% Because the severity distribution is highly imbalanced, the discussion emphasizes Macro F1 and balanced accuracy; accuracy and macro precision are also reported for completeness. MCC complements the class-balanced metrics by summarizing agreement over the complete multiclass confusion matrix. This emphasis is necessary because overall accuracy may overestimate the effectiveness of a model that mainly captures the dominant low-severity class.

\subsection{Multilabel CODE prediction effectiveness} 
\label{subsec5.1} 

\nd The first stage of the proposed framework predicts defect CODE degrees from sewer pipe inspection images. Table~\ref{tab:multilabel_code_results} reports the effectiveness of the multilabel CODE prediction model and compares it with a random multilabel baseline. The Swin Transformer-based multilabel model (Swin-ML) substantially outperforms the random baseline across all metrics. 
% In particular, the proposed multilabel model achieves an exact match accuracy of 0.5920, a Hamming accuracy of 0.9493, a Macro F1 score of 0.4689, and a mean average precision of 0.6197.
% Define 'L' as a left-aligned column with a specific width (e.g., 20% of text width)
\newcolumntype{L}{>{\raggedright\arraybackslash}p{0.2\textwidth}}
\begin{table}[t]
\caption{Effectiveness of image-based multilabel CODE prediction on the test set.}
\label{tab:multilabel_code_results}
\centering
\setlength{\tabcolsep}{4pt}
\begin{tabular*}{\textwidth}{@{\extracolsep{\fill}}lcccc@{}}
\toprule
Method & 
\shortstack{Exact match\\accuracy} & 
\shortstack{Hamming\\accuracy} & 
\shortstack{Macro\\F1} & 
mAP \\
\midrule
Random & 0.1610 & 0.8779 & 0.0722 & 0.0722 \\
Swin-ML & 0.6963 & 0.9748 & 0.4116 & 0.5616 \\
\bottomrule
\end{tabular*}
\end{table}
The high Hamming accuracy indicates that the model correctly predicts most label-wise CODE decisions. However, this metric should be interpreted with caution because the multilabel CODE space is sparse and most CODE labels are absent in most images. Macro F1 and mAP are therefore more informative for assessing the ability of the model to recognize less frequent defect CODEs. The obtained Macro F1 score indicates that several CODE labels remain difficult to predict, especially those that are visually subtle, rare, or affected by image noise.

These results have direct implications for the subsequent symbolic reasoning stage.  The rule layer depends on the quality of the predicted CODE degrees. If a relevant CODE is missed or an irrelevant CODE receives a high degree, the fixed rule base may receive an incomplete or misleading semantic representation. 

% Nevertheless, the multilabel predictor provides a meaningful intermediate interface between raw visual information and symbolic severity reasoning. Instead of forcing the image backbone to directly predict severity, the proposed framework first maps the inspection image into defect-related semantic evidence.

\subsection{Overall severity prediction effectiveness} 
\label{subsec5.2} 

\nd Table~\ref{tab:severity_main_results} compares the proposed framework with the image-only baseline and the Oracle CODE + Rule setting\footnote{For completeness, the confusion matrixes are shown in Appendix~\ref{app2}.}. The image-only baseline directly predicts the severity class from the inspection image. The proposed Multilabel + Rule framework first predicts CODE degrees and then performs fuzzy rule-based severity inference using the fixed symbolic rule base. For each fuzzy operator, the interface mode was selected on the validation set from the soft-degree configuration and nine thresholds. Interestingly, this procedure selected soft-degree inference for all proposed fuzzy operators (P, L, and H). The Oracle CODE + Rule setting replaces image-predicted CODE degrees with ground-truth CODE annotations and is included only as a diagnostic upper-bound experiment.

\begin{table*}[t]
\caption{Severity prediction effectiveness on the test set. The interface mode of each proposed variant was selected exclusively on the validation set using Macro F1.}
\label{tab:severity_main_results}
\centering
\small
\setlength{\tabcolsep}{4pt}
\begin{tabular*}{\textwidth}{@{\extracolsep{\fill}}p{0.15\textwidth}ccccc@{}}
\toprule
Method & (Accuracy) & Balanced Accuracy & Macro Precision & Macro F1 & MCC\\
\midrule
Image-only severity baseline 
& 0.6350 & 0.4697 & 0.4482 & 0.4350 & 0.4176\\\\

Proposed, Product (soft degree)
& 0.7485 & 0.5272 & 0.7172 & 0.5352 & 0.4901 \\\\

Proposed, {\L}ukasiewicz (soft degree)
& 0.7485 & 0.5272 & 0.6974 & 0.5220 & 0.4914\\\\

Proposed, Hamacher (soft degree)
& 0.7485 & 0.5272 & 0.7172 & 0.5352 & 0.4901\\\\

Oracle CODE + Rule, Product 
& 0.8620 & 0.7339 & 0.8097 & 0.7263 & 0.7351\\\\
\bottomrule
\end{tabular*}
\end{table*}

\nd The image-only baseline obtains a Macro F1 score of 0.4350, a balanced accuracy of 0.4697, and an MCC of 0.4176. This confirms that direct severity classification from images is difficult under the present data distribution. Although the model can learn visual patterns associated with frequent severity categories, it has limited ability to explicitly model the defect-level reasoning process that links visual evidence to severity decisions.

The Product and Hamacher variants attain an accuracy of 0.7485, a balanced accuracy of 0.5272, a Macro F1 of 0.5352, and an MCC of 0.4901. Relative to the image-only baseline, these values correspond to improvements of approximately 17.9\%, 12.2\%, 23.0\%, and 17.3\%, respectively. Macro precision increases from 0.4482 to 0.7172 improving 60\%. Product and Hamacher produce identical aggregate test metrics under soft-degree inference.

The {\L}ukasiewicz soft-degree variant has the same accuracy and balanced accuracy, a slightly lower Macro F1 of 0.5220, but the highest deployable MCC of 0.4914 (+17.7\%). Thus, Product and Hamacher provide the strongest class-wise F1 balance, whereas {\L}ukasiewicz provides a marginally higher global agreement. All three DT-based variants outperform the image-only baseline across the reported aggregate metrics.

The Oracle CODE + Rule setting obtains the strongest results, with a balanced accuracy of 0.7339, a Macro F1 of 0.7263, and an MCC of 0.7351. Its MCC exceeds that of the Product and Hamacher variants by approximately 50.0\%, showing that reliable ground-truth CODE inputs substantially increase agreement over the complete confusion matrix.
We recall that the Oracle CODE + Rule experiment provides a diagnostic upper-bound analysis of the symbolic reasoning module. In this setting, ground-truth CODE annotations are directly passed to the fixed rule base and, thus, should not be interpreted as deployable image-only performance. Since this configuration does not depend on image-based CODE prediction, it evaluates whether the extracted rules can map CODE patterns to severity labels when the semantic input representation is reliable. Nevertheless, the non-negligible improvement confirms that the extracted rule base contains useful severity knowledge and can support effective severity inference when CODE inputs are accurate.
At the same time, the gap between the Oracle setting and the image-predicted CODE settings identifies CODE recognition as a major source of downstream error. The Oracle result is still below perfect effectiveness, indicating that the symbolic reasoning layer is not the only remaining limitation. Rule coverage, severity-label uncertainty, class imbalance, and overlap between CODE patterns associated with different severity classes may also contribute to errors. Improving CODE recognition, particularly for rare or visually ambiguous defect CODEs, is nevertheless expected to improve definitely the final severity assessment.

% The Oracle result should not be interpreted as deployable image-only performance because ground-truth CODE annotations are not available during inference. Instead, it separates the potential of the symbolic reasoning module from the limitations of the neural CODE prediction stage.

% Because this setting uses ground-truth CODE annotations, it is not deployable from images alone. 
% Instead, it estimates the effectiveness of the fixed rule layer when the semantic CODE representation is reliable. 
% The gap between the Oracle and image-predicted CODE settings identifies CODE recognition as a major source of downstream error, while the remaining Oracle errors also indicate limitations in rule coverage, label quality, or the separability of the severity classes.

%\todo[inline]{UMBERTO: I've read till  HERE}

\subsection{Validation-based threshold selection and sensitivity analysis}
\label{subsec5.3} 

\nd The neuro-symbolic interface was examined in two modes. In soft-degree mode, predicted CODE degrees are passed directly to the fuzzy rule layer. In threshold mode, each degree is converted into a binary value before rule evaluation. For each operator, these modes form a common set of ten validation candidates: one soft-degree configuration and nine thresholds in $\{0.1,0.2,\ldots,0.9\}$. All ten candidates are ranked using validation Macro F1, without access to the test labels.

Table~\ref{tab:threshold_selection_validation} reports the validation comparison. Soft-degree results are operator-specific because the t-norm and s-norm pairs operate directly on continuous CODE degrees and confidence-weighted rule supports. In threshold mode, all operators receive the same binary CODE inputs and produce identical aggregate validation metrics; these rows are therefore reported once. Among the thresholded candidates, $\tau=0.5$ attains the highest Macro F1 (0.5276).

\begin{table*}[t]
\caption{Validation-set comparison of the ten CODE-interface candidates considered for each fuzzy operator. Thresholded results are identical across operators and are reported once. Bold rows identify the operator-specific selected configurations.}
\label{tab:threshold_selection_validation}
\centering
\scriptsize
\setlength{\tabcolsep}{4pt}
\begin{tabular}{lllcccp{2.0cm}}
\toprule
Interface & Operator & $\tau$ & Accuracy & Balanced Acc. &
Macro F1  & Selected for \\
\midrule
\textbf{Soft degree} & \textbf{Product} & \textbf{--} & \textbf{0.7362} & \textbf{0.4959} & \textbf{0.5309} & \textbf{Product} \\
\textbf{Soft degree} & \textbf{Hamacher} & \textbf{--} & \textbf{0.7362} & \textbf{0.4959} & \textbf{0.5309} & \textbf{Hamacher} \\
\textbf{Soft degree} & \textbf{{\L}ukasiewicz} & \textbf{--} & \textbf{0.7331} & \textbf{0.4949} & \textbf{0.5291} & \textbf{{\L}ukasiewicz} \\
\midrule
Threshold & All & 0.1 & 0.6963 & 0.5221 & 0.5112 & -- \\
Threshold & All & 0.2 & 0.7025 & 0.5143 & 0.5190 & -- \\
Threshold & All & 0.3 & 0.7147 & 0.4999 & 0.5153 & -- \\
Threshold & All & 0.4 & 0.7209 & 0.5018 & 0.5260 & -- \\
Threshold & All & 0.5 & 0.7301 & 0.4948 & 0.5276 & -- \\
Threshold & All & 0.6 & 0.7270 & 0.4820 & 0.5189 & -- \\
Threshold & All & 0.7 & 0.7301 & 0.4634 & 0.5012 & -- \\
Threshold & All & 0.8 & 0.7209 & 0.4190 & 0.4467 & -- \\
Threshold & All & 0.9 & 0.7331 & 0.4008 & 0.4337 & -- \\
\bottomrule
\end{tabular}
\end{table*}

Product and Hamacher attain a validation Macro F1 of 0.5309 in soft-degree mode, exceeding the best thresholded result by 0.0033. {\L}ukasiewicz attains a soft-degree Macro F1 of 0.5291, exceeding the best thresholded result by 0.0015. Consequently, soft-degree inference is selected for all three operators. These small margins indicate that the selected interface may remain configuration- and split-dependent.

Within the threshold configurations, different criteria favour different thresholds. The threshold $\tau=0.9$ gives the highest accuracy, $\tau=0.1$ gives the highest balanced accuracy, and $\tau=0.5$ gives the highest Macro F1. None is selected because soft-degree inference yields the highest Macro F1 for every operator. The deterioration in balanced accuracy at high thresholds is consistent with strict binarization suppressing relevant CODE evidence before rule evaluation.

The three selected configurations are fixed at this stage and subsequently evaluated once on the test set, as reported in Table~\ref{tab:severity_main_results}. Restricting the main test analysis to validation-selected configurations preserves the separation between model selection and final evaluation.

\subsection{Qualitative explanation analysis} 
\label{subsec5.5} 

\nd Beyond aggregate effectiveness, the proposed framework provides an explicit numerical trace from the predicted CODE degrees to the final severity decision. The analysis uses the validation-selected Product soft-degree configuration evaluated in Table~\ref{tab:severity_main_results}. No threshold is applied: positive and negative antecedents are evaluated as $\mu_{ij}$ and $1-\mu_{ij}$, respectively, before Product activation, confidence weighting, and Probabilistic Sum aggregation.

For each of the 326 test samples, the exported trace contains all 14 predicted CODE degrees and the activation and support values of all 19 rules. Because several rules may receive non-zero support under soft-degree inference, the representative rule for sample $i$ is defined as the highest-supported rule among those concluding the predicted class:
\[
r_i^\star=
\arg\max_{r\in\mathcal{R}_{\hat{y}_i}}
\operatorname{Supp}_{ir}^{(\mathrm{P})}.
\]
This rule identifies the strongest individual contribution to the predicted class; the final decision still depends on the aggregation of all rule supports within each class. To reduce subjective example selection, a fixed-seed stratified random protocol (seed 42) was used. One correctly predicted sample was drawn from each severity class, whose eligible pools contained 196, 27, 17, 1, and 3 samples for classes 1--5, respectively. In addition, one class-4 sample misclassified as class 3 was drawn from the corresponding pool of four cases to examine the class-3/class-4 boundary. For interpretation, the evidence margin of each selected case is reported as
\[
\Delta_i =
E_i(\hat{y}_i)-\max_{c\neq\hat{y}_i}E_i(c).
\]
The exported traces were numerically checked by recomputing the Product class evidence from all rule supports. The predicted classes were identical and the maximum evidence discrepancy was $9.65\times10^{-13}$.

%\begin{table*}[t]
%\caption{Numerical inference traces for the selected test cases under the validation-selected Product soft-degree configuration. The class-evidence vector is ordered from severity 1 to 5, and $r_i^\star$ denotes the highest-supported rule for the predicted class.}
%\label{tab:qualitative_examples}
%\centering
%\scriptsize
%\setlength{\tabcolsep}{7pt}
%\begin{tabular}{@{}lccccccp{5.4cm}c@{}}
%\toprule
%Case (image) & True & Pred. & $r_i^\star$ & $A_{ir_i^\star}$ &
%$\operatorname{Conf}_{r_i^\star}$ & $\operatorname{Supp}_{ir_i^\star}$ &
%$\big(E_i(1),E_i(2),E_i(3),E_i(4),E_i(5)\big)$ & $\Delta_i$ \\
%\midrule
%C1 (18.jpg)  & 1 & 1 & $R_1$    & 0.997949 & 0.884989 & 0.883174 &
%$(0.883178,\,0.000088,\,0.001354,\,1.20{\times}10^{-8},\,0.000014)$ & 0.881824 \\
%C2 (11.jpg)  & 2 & 2 & $R_{15}$ & 0.992502 & 0.879310 & 0.872717 &
%$(0.001921,\,0.873289,\,0.000079,\,2.98{\times}10^{-9},\,0.000107)$ & 0.871368 \\
%C3 (59.jpg)  & 3 & 3 & $R_{16}$ & 0.981844 & 0.712821 & 0.699879 &
%$(0.014590,\,0.000062,\,0.700156,\,0.000222,\,0.000045)$ & 0.685566 \\
%C4 (57.jpg)  & 4 & 4 & $R_{13}$ & 0.723621 & 0.800000 & 0.578897 &
%$(0.000324,\,0.000149,\,0.059851,\,0.578964,\,0.098085)$ & 0.480879 \\
%C5 (43.jpg)  & 5 & 5 & $R_{19}$ & 0.779746 & 0.610169 & 0.475777 &
%$(0.139960,\,0.000025,\,0.046246,\,0.000002,\,0.475777)$ & 0.335817 \\
%C6 (113.jpg) & 4 & 3 & $R_{12}$ & 0.685692 & 0.520000 & 0.356560 &
%$(0.086536,\,0.008897,\,0.358356,\,0.001614,\,0.122497)$ & 0.235860 \\
%\bottomrule
%\end{tabular}
%\end{table*}

\begin{table*}[t]
\caption{Numerical inference traces for the selected test cases under the validation-selected Product soft-degree configuration. The class-evidence vector is ordered from severity 1 to 5, and $r_i^\star$ denotes the highest-supported rule for the predicted class.}
\label{tab:qualitative_examples}
\centering
% \scriptsize and \tabcolsep removed because resizebox handles the sizing automatically
\resizebox{\textwidth}{!}{%
\begin{tabular}{@{}lcccccccc@{}} % Changed p{5.4cm} to c
\toprule
Case (image) & True & Pred. & $r_i^\star$ & $A_{ir_i^\star}$ &
$\operatorname{Conf}_{r_i^\star}$ & $\operatorname{Supp}_{ir_i^\star}$ &
$\big(E_i(1),E_i(2),E_i(3),E_i(4),E_i(5)\big)$ & $\Delta_i$ \\
\midrule
C1 (18.jpg)  & 1 & 1 & $R_1$    & 0.997949 & 0.884989 & 0.883174 &
$(0.883178,\,0.000088,\,0.001354,\,1.20{\times}10^{-8},\,0.000014)$ & 0.881824 \\
C2 (11.jpg)  & 2 & 2 & $R_{15}$ & 0.992502 & 0.879310 & 0.872717 &
$(0.001921,\,0.873289,\,0.000079,\,2.98{\times}10^{-9},\,0.000107)$ & 0.871368 \\
C3 (59.jpg)  & 3 & 3 & $R_{16}$ & 0.981844 & 0.712821 & 0.699879 &
$(0.014590,\,0.000062,\,0.700156,\,0.000222,\,0.000045)$ & 0.685566 \\
C4 (57.jpg)  & 4 & 4 & $R_{13}$ & 0.723621 & 0.800000 & 0.578897 &
$(0.000324,\,0.000149,\,0.059851,\,0.578964,\,0.098085)$ & 0.480879 \\
C5 (43.jpg)  & 5 & 5 & $R_{19}$ & 0.779746 & 0.610169 & 0.475777 &
$(0.139960,\,0.000025,\,0.046246,\,0.000002,\,0.475777)$ & 0.335817 \\
C6 (113.jpg) & 4 & 3 & $R_{12}$ & 0.685692 & 0.520000 & 0.356560 &
$(0.086536,\,0.008897,\,0.358356,\,0.001614,\,0.122497)$ & 0.235860 \\
\bottomrule
\end{tabular}%
}
\end{table*}

Table~\ref{tab:qualitative_antecedents} reports the predicted CODE degrees entering each dominant rule. The superscripts $+$ and $-$ denote presence and absence predicates, respectively, while the bracketed value is the corresponding predicted CODE degree $\mu_{ij}$. Thus, the condition degree is the bracketed value for a $+$ predicate and its complement for a $-$ predicate.

\begin{table*}[t]
\caption{Antecedent-level traces for the highest-supported predicted-class rules in Table~\ref{tab:qualitative_examples}. Bracketed values are predicted CODE degrees before complementing absence predicates and may refer to different images.}
\label{tab:qualitative_antecedents}
\centering
\scriptsize
\setlength{\tabcolsep}{4pt}
% Replace tabular with tabularx, set total width to \textwidth, change p{14.2cm} to X
\begin{tabularx}{\textwidth}{@{}lcX@{}} 
\toprule
Case & Rule & Antecedent predicates and predicted CODE degrees \\
\midrule
C1 & $R_1$ &
\(\mathrm{BAC}^{-}[0.000024]\), \(\mathrm{BDD}^{-}[0.000027]\),
\(\mathrm{BAA}^{-}[0.000107]\), \(\mathrm{BAF}^{-}[0.000072]\),
\(\mathrm{BAB}^{-}[0.000156]\), \(\mathrm{BBA}^{-}[0.000109]\),
\(\mathrm{BDC}^{-}[0.000035]\), \(\mathrm{BCE}^{-}[0.000014]\),
\(\mathrm{BAI}^{-}[0.000391]\), \(\mathrm{BAJ}^{-}[0.001065]\),
\(\mathrm{BBC}^{-}[0.000020]\), \(\mathrm{BBB}^{-}[0.000034]\) \\
C2 & $R_{15}$ &
\(\mathrm{BAC}^{-}[0.000175]\), \(\mathrm{BDD}^{-}[0.005044]\),
\(\mathrm{BAA}^{-}[0.000097]\), \(\mathrm{BAF}^{+}[0.997804]\) \\
C3 & $R_{16}$ &
\(\mathrm{BAC}^{-}[0.000074]\), \(\mathrm{BDD}^{-}[0.000068]\),
\(\mathrm{BAA}^{+}[0.982206]\), \(\mathrm{BDC}^{-}[0.000226]\) \\
C4 & $R_{13}$ &
\(\mathrm{BAC}^{-}[0.160751]\), \(\mathrm{BDD}^{-}[0.000120]\),
\(\mathrm{BAA}^{-}[0.000221]\), \(\mathrm{BAF}^{-}[0.000081]\),
\(\mathrm{BAB}^{-}[0.000595]\), \(\mathrm{BBA}^{+}[0.999452]\),
\(\mathrm{BDC}^{+}[0.863575]\) \\
C5 & $R_{19}$ &
\(\mathrm{BAC}^{+}[0.779746]\) \\
C6 & $R_{12}$ &
\(\mathrm{BAC}^{-}[0.200759]\), \(\mathrm{BDD}^{-}[0.011717]\),
\(\mathrm{BAA}^{-}[0.002879]\), \(\mathrm{BAF}^{-}[0.000799]\),
\(\mathrm{BAB}^{-}[0.001431]\), \(\mathrm{BBA}^{+}[0.875107]\),
\(\mathrm{BDC}^{-}[0.002921]\) \\
\bottomrule
\end{tabularx}
\end{table*}

Cases C1--C5 illustrate correct predictions through distinct rules. C1 activates the long absence-pattern rule $R_1$, producing class-1 evidence of 0.883178. C2 and C3 are dominated by the high BAF and BAA degrees in $R_{15}$ and $R_{16}$, respectively. For C5, the BAC degree of 0.779746 directly activates the single-antecedent class-5 rule $R_{19}$, producing class-5 evidence of 0.475777.

The pair C4--C6 exposes the class-3/class-4 boundary. Rules $R_{12}$ and $R_{13}$ share the same preceding antecedents, including \(\mathrm{BBA}^{+}\), but use complementary BDC conditions: $R_{12}$ concludes class 3 when BDC is absent, whereas $R_{13}$ concludes class 4 when BDC is present. For the correctly classified C4 case, the BDC degree is 0.863575 and $R_{13}$ contributes 0.578897 to class 4. For C6, the BDC degree is 0.002921; its complement strongly supports $R_{12}$, yielding class-3 evidence of 0.358356, while class-4 evidence remains 0.001614. This trace identifies the low predicted BDC degree as the immediate computational reason for the class-4-to-class-3 error and clarifies why errors between these classes can arise under the rule structure.

Overall, the cases show that both correct and incorrect predictions can be decomposed into predicted CODE degrees, antecedent degrees, rule activations, confidence-weighted supports, and aggregated class evidence under the same validation-selected configuration used for quantitative evaluation. The analysis demonstrates traceability of individual decisions, while the aggregate test results remain the basis for assessing predictive effectiveness.

\subsection{Discussion and limitations} 
\label{subsec5.6} 

\nd The experimental results show that using defect CODEs as an intermediate semantic representation improves the interpretability and class-balanced effectiveness of image-based severity assessment. The proposed framework decomposes severity prediction into neural CODE prediction and symbolic rule reasoning, in contrast to the image-only baseline, which directly maps images to severity labels.

The comparison in Table~\ref{tab:severity_main_results} shows that our method outperforms the image-only baseline across all reported metrics. For Product and Hamacher, relative gains reach 17.9\% in accuracy, 12.2\% in balanced accuracy, 23.0\% in Macro F1, and 17.3\% in MCC. The MCC increase confirms that the improvement extends beyond the two class-balanced metrics. Nevertheless, the class-wise analysis shows that these aggregate gains coexist with weak class-4 recall and should not be interpreted as uniform improvement across severity levels.

The validation analysis selects soft-degree inference for Product, Hamacher, and {\L}ukasiewicz. Product and Hamacher yield identical aggregate test metrics under this interface, but aggregate equality alone does not establish identical rule activations or class evidence. {\L}ukasiewicz produces the same accuracy and balanced accuracy but differs slightly in Macro F1, Macro Precision, MCC, and several confusion-matrix cells.

The validation results also indicate that interface selection may be split-dependent. The soft-degree advantage over the best thresholded candidate is only 0.0033 Macro F1 for Product and Hamacher and 0.0015 for {\L}ukasiewicz. These differences should be interpreted as configuration-specific findings rather than evidence that one interface will generalize best to other datasets or coding standards.

The Oracle CODE + Rule experiment clarifies the role of the symbolic rule base. Its strong effectiveness, including an MCC of 0.7341, indicates that the rule base can map CODE patterns to severity classes when reliable CODE inputs are available. The approximately 50.0\% MCC gap relative to Product and Hamacher identifies multilabel CODE prediction as an important bottleneck, consistent with the moderate CODE-level Macro F1 despite high Hamming accuracy. Because Oracle reasoning is not perfect, however, rule coverage and label uncertainty also remain relevant.

Several limitations remain. First, the framework depends on the quality of predicted CODE degrees. If critical CODEs are missed, the rule layer may activate an incorrect severity rule. Second, the selected operator--interface configurations are based on one validation split; the small differences between several candidates suggest that their stability should be examined using repeated splits or cross-validation when sufficient data are available. Third, the rule base is extracted from a single DT, which keeps the reasoning trace compact but may limit coverage for rare defect combinations. 
% Fourth, the current evaluation uses image-level splitting and does not enforce source-file-level separation. Future work should therefore consider stricter inspection-session-level or pipe-segment-level evaluation protocols when such metadata are available. 
Finally, the severity labels are generated using a multi-LLM consensus protocol based on inspector-written observations; additional expert validation would further strengthen the reliability of the evaluation.

Overall, the results support the proposed framework as an interpretable image-based severity assessment approach. Among the deployable methods, the framework achieves the strongest results across all reported metrics, including balanced accuracy, Macro F1, and MCC, while providing explicit rule-level explanations. The Oracle analysis further shows substantial diagnostic potential when ground-truth CODE annotations are available. The remaining effectiveness gap points primarily, but not exclusively, to the need for stronger multilabel CODE prediction.

\section{Conclusion}
\label{sec6}

\nd This study presented a fuzzy rule-based neuro-symbolic framework for image-based sewer pipe severity assessment. The framework decomposes the task into neural multilabel prediction of 14 defect CODE degrees and fuzzy severity reasoning over a fixed symbolic rule base extracted via a DT (viz.~Weka's J48 algorithm). This design preserves an image-only deployment setting while exposing the intermediate CODE representation, activated IF--THEN rules, rule confidence values, and aggregated class evidence underlying each severity prediction. Product, {\L}ukasiewicz, and Hamacher t-norm/s-norm pairs were evaluated together with soft-degree and thresholded CODE interfaces, with the interface configuration selected exclusively on the validation set.

The validation-selected Product and Hamacher soft-degree configurations achieved the highest deployable Macro F1, while {\L}ukasiewicz achieved a marginally higher MCC than the two before. All three variants outperformed the image-only baseline across the reported aggregate metrics. The Oracle CODE + Rule experiment indicated that the fixed rule base contains useful severity knowledge and that imperfect CODE prediction is a major, although not the only, source of downstream error. The qualitative cases further showed that both correct and incorrect decisions can be examined through the same explicit rule-level inference trace.

% The results also indicate that the fuzzy operator and the neuro-symbolic interface should be considered jointly: soft-degree inference was selected for Product and Hamacher, whereas thresholding at $\tau=0.4$ was selected for {\L}ukasiewicz. These findings seem to be specific to the adopted validation split and do not establish a universal preference for either continuous or threshold CODE inputs. Further work should improve recognition of rare and visually ambiguous CODEs, examine the stability of operator--interface selection under repeated or cross-validated evaluation, and investigate richer but still interpretable rule bases. Also, domain-expert validation of the multi-LLM-derived severity labels may strengthen the reliability of the target annotations. Last, but not least, other rule extraction methods may be investigated, such as RIPPER~\cite{Cohen95} and J48 (see~\cite{Quinlan93,Witten11}). Addressing these limitations may improve predictive robustness while retaining the transparent separation between visual perception and symbolic severity reasoning.

Soft-degree inference was selected for all three fuzzy operators, although its validation advantage over the best thresholded candidate was small. Further work should improve recognition of rare and visually ambiguous CODEs, examine the stability of interface selection under repeated or cross-validated evaluation, and investigate richer but still interpretable rule extractors, such as RIPPER~\cite{Cohen95}. Domain-expert validation of the multi-LLM-derived severity labels may also strengthen the reliability of the target annotations.

\section*{Acknowledgements}
\nd This research has received support from the European Union's Horizon research and innovation program (under the MSCA-SE (Marie Sk\l{}odowska-Curie Actions Staff Exchange) grant agreement 101086252; Call: HORIZON-MSCA-2021-SE-01; Project title: STARWARS (STormwAteR and WastewAteR networkS heterogeneous data AI-driven management).

\appendix
\section{Appendix: Attributes and Rule Base} \label{app1}

\nd In the following, we summarize the automatically extracted diagnostic rules from the training dataset.  
% Let $X$ denote a dermoscopic image, $D$ a fuzzy activation degree, $O_A$ the selected fuzzy AND operator, and $O_R$ the selected fuzzy OR operator. Each attribute predicate returns the predicted fuzzy membership degree of the corresponding dermoscopic attribute category for image $X$. The fuzzy operator equations are defined in the main text; therefore, only their semantic representation is provided here.

%\vspace*{2em}
%\nd {\bf \large Attribute membership predicates} \mbox{ \\ }
\subsection*{CODEs and Attribute Predicates} 
%\vspace*{0.5cm}

\nd Let $X$ be a variable denoting a sewer pipe inspection image. For each retained inspection CODE $\mathtt{codej}$, the truth degree of the attribute predicate
\[
\mathtt{codej\_present}(X)\]
\nd is given by the predicted CODE degree $\mu_j(X)$, whereas the truth degree of the attribute predicate
\[
\mathtt{codej\_absent}(X)
\]
\nd is its complement $1-\mu_j(X)$.

%\vspace*{0.3cm}

The CODEs and attribute predicates of the rule base are the following:

\begin{center}
%\scriptsize
\footnotesize
\renewcommand{\arraystretch}{1.15}
\begin{tabular}{@{}p{0.40\columnwidth}p{0.52\columnwidth}@{}}
\hline
Sewer inspection CODEs & Attribute predicates \\ \hline

Deformation (BAA) &
$\mathtt{baa\_absent}(X)$,
$\mathtt{baa\_present}(X)$ \\

Crack/fissure (BAB) &
$\mathtt{bab\_absent}(X)$,
$\mathtt{bab\_present}(X)$ \\

Break/collapse (BAC) &
$\mathtt{bac\_absent}(X)$,
$\mathtt{bac\_present}(X)$ \\

Surface damage (BAF) &
$\mathtt{baf\_absent}(X)$,
$\mathtt{baf\_present}(X)$ \\

Intruding connection (BAG) &
$\mathtt{bag\_absent}(X)$,
$\mathtt{bag\_present}(X)$ \\

Intruding sealing material (BAI) &
$\mathtt{bai\_absent}(X)$,
$\mathtt{bai\_present}(X)$ \\

Displaced joint (BAJ) &
$\mathtt{baj\_absent}(X)$,
$\mathtt{baj\_present}(X)$ \\

Roots in pipeline (BBA) &
$\mathtt{bba\_absent}(X)$,
$\mathtt{bba\_present}(X)$ \\

Attached deposits (BBB) &
$\mathtt{bbb\_absent}(X)$,
$\mathtt{bbb\_present}(X)$ \\

Settled deposits (BBC) &
$\mathtt{bbc\_absent}(X)$,
$\mathtt{bbc\_present}(X)$ \\

Finish node (BCE) &
$\mathtt{bce\_absent}(X)$,
$\mathtt{bce\_present}(X)$ \\

Inspection abandoned (BDC) &
$\mathtt{bdc\_absent}(X)$,
$\mathtt{bdc\_present}(X)$ \\

Water level (BDD) &
$\mathtt{bdd\_absent}(X)$,
$\mathtt{bdd\_present}(X)$ \\

Roots in manhole or inspection chamber (DBA) &
$\mathtt{dba\_absent}(X)$,
$\mathtt{dba\_present}(X)$ \\

\hline
\end{tabular}
\end{center}

%\vspace*{2em}
%\nd {\bf \large Diagnostic rules} \mbox{ \\ }
\subsection*{Rule-base} \mbox{ \\ }
%\vspace*{2em}
\nd The extracted rule base is the following:

{%
\footnotesize

\[
\begin{array}{llcl}

R_1: &
\mathtt{sev\_1}(X) &
\leftarrow &
\mathtt{bac\_absent}(X), \mathtt{bdd\_absent}(X), \\
&&&
\mathtt{baa\_absent}(X), \mathtt{baf\_absent}(X), \\
&&&
\mathtt{bab\_absent}(X), \mathtt{bba\_absent}(X), \\
&&&
\mathtt{bdc\_absent}(X), \mathtt{bce\_absent}(X), \\
&&&
\mathtt{bai\_absent}(X), \mathtt{baj\_absent}(X), \\
&&&
\mathtt{bbc\_absent}(X), \mathtt{bbb\_absent}(X)
\\[1.1em]

R_2: &
\mathtt{sev\_3}(X) &
\leftarrow &
\mathtt{bac\_absent}(X), \mathtt{bdd\_absent}(X), \\
&&&
\mathtt{baa\_absent}(X), \mathtt{baf\_absent}(X), \\
&&&
\mathtt{bab\_absent}(X), \mathtt{bba\_absent}(X), \\
&&&
\mathtt{bdc\_absent}(X), \mathtt{bce\_absent}(X), \\
&&&
\mathtt{bai\_absent}(X), \mathtt{baj\_absent}(X), \\
&&&
\mathtt{bbc\_absent}(X), \mathtt{bbb\_present}(X)
\\[1.1em]

R_3: &
\mathtt{sev\_3}(X) &
\leftarrow &
\mathtt{bac\_absent}(X), \mathtt{bdd\_absent}(X), \\
&&&
\mathtt{baa\_absent}(X), \mathtt{baf\_absent}(X), \\
&&&
\mathtt{bab\_absent}(X), \mathtt{bba\_absent}(X), \\
&&&
\mathtt{bdc\_absent}(X), \mathtt{bce\_absent}(X), \\
&&&
\mathtt{bai\_absent}(X), \mathtt{baj\_absent}(X), \\
&&&
\mathtt{bbc\_present}(X)
\\[1.1em]

R_4: &
\mathtt{sev\_3}(X) &
\leftarrow &
\mathtt{bac\_absent}(X), \mathtt{bdd\_absent}(X), \\
&&&
\mathtt{baa\_absent}(X), \mathtt{baf\_absent}(X), \\
&&&
\mathtt{bab\_absent}(X), \mathtt{bba\_absent}(X), \\
&&&
\mathtt{bdc\_absent}(X), \mathtt{bce\_absent}(X), \\
&&&
\mathtt{bai\_absent}(X), \mathtt{baj\_present}(X)
\\[1.1em]

R_5: &
\mathtt{sev\_3}(X) &
\leftarrow &
\mathtt{bac\_absent}(X), \mathtt{bdd\_absent}(X), \\
&&&
\mathtt{baa\_absent}(X), \mathtt{baf\_absent}(X), \\
&&&
\mathtt{bab\_absent}(X), \mathtt{bba\_absent}(X), \\
&&&
\mathtt{bdc\_absent}(X), \mathtt{bce\_absent}(X), \\
&&&
\mathtt{bai\_present}(X)

\end{array}
\]

\[
\begin{array}{llcl}

R_6: &
\mathtt{sev\_1}(X) &
\leftarrow &
\mathtt{bac\_absent}(X), \mathtt{bdd\_absent}(X), \\
&&&
\mathtt{baa\_absent}(X), \mathtt{baf\_absent}(X), \\
&&&
\mathtt{bab\_absent}(X), \mathtt{bba\_absent}(X), \\
&&&
\mathtt{bdc\_absent}(X), \mathtt{bce\_present}(X), \\
&&&
\mathtt{dba\_absent}(X)
\\[1.1em]

R_7: &
\mathtt{sev\_3}(X) &
\leftarrow &
\mathtt{bac\_absent}(X), \mathtt{bdd\_absent}(X), \\
&&&
\mathtt{baa\_absent}(X), \mathtt{baf\_absent}(X), \\
&&&
\mathtt{bab\_absent}(X), \mathtt{bba\_absent}(X), \\
&&&
\mathtt{bdc\_absent}(X), \mathtt{bce\_present}(X), \\
&&&
\mathtt{dba\_present}(X)
\\[1.1em]

R_8: &
\mathtt{sev\_1}(X) &
\leftarrow &
\mathtt{bac\_absent}(X), \mathtt{bdd\_absent}(X), \\
&&&
\mathtt{baa\_absent}(X), \mathtt{baf\_absent}(X), \\
&&&
\mathtt{bab\_absent}(X), \mathtt{bba\_absent}(X), \\
&&&
\mathtt{bdc\_present}(X), \mathtt{bag\_absent}(X), \\
&&&
\mathtt{bbb\_absent}(X), \mathtt{bbc\_absent}(X)
\\[1.1em]

R_9: &
\mathtt{sev\_4}(X) &
\leftarrow &
\mathtt{bac\_absent}(X), \mathtt{bdd\_absent}(X), \\
&&&
\mathtt{baa\_absent}(X), \mathtt{baf\_absent}(X), \\
&&&
\mathtt{bab\_absent}(X), \mathtt{bba\_absent}(X), \\
&&&
\mathtt{bdc\_present}(X), \mathtt{bag\_absent}(X), \\
&&&
\mathtt{bbb\_absent}(X), \mathtt{bbc\_present}(X)
\\[1.1em]

R_{10}: &
\mathtt{sev\_4}(X) &
\leftarrow &
\mathtt{bac\_absent}(X), \mathtt{bdd\_absent}(X), \\
&&&
\mathtt{baa\_absent}(X), \mathtt{baf\_absent}(X), \\
&&&
\mathtt{bab\_absent}(X), \mathtt{bba\_absent}(X), \\
&&&
\mathtt{bdc\_present}(X), \mathtt{bag\_absent}(X), \\
&&&
\mathtt{bbb\_present}(X)

\end{array}
\]

\[
\begin{array}{llcl}

R_{11}: &
\mathtt{sev\_4}(X) &
\leftarrow &
\mathtt{bac\_absent}(X), \mathtt{bdd\_absent}(X), \\
&&&
\mathtt{baa\_absent}(X), \mathtt{baf\_absent}(X), \\
&&&
\mathtt{bab\_absent}(X), \mathtt{bba\_absent}(X), \\
&&&
\mathtt{bdc\_present}(X), \mathtt{bag\_present}(X)
\\[1.1em]

R_{12}: &
\mathtt{sev\_3}(X) &
\leftarrow &
\mathtt{bac\_absent}(X), \mathtt{bdd\_absent}(X), \\
&&&
\mathtt{baa\_absent}(X), \mathtt{baf\_absent}(X), \\
&&&
\mathtt{bab\_absent}(X), \mathtt{bba\_present}(X), \\
&&&
\mathtt{bdc\_absent}(X)
\\[1.1em]

R_{13}: &
\mathtt{sev\_4}(X) &
\leftarrow &
\mathtt{bac\_absent}(X), \mathtt{bdd\_absent}(X), \\
&&&
\mathtt{baa\_absent}(X), \mathtt{baf\_absent}(X), \\
&&&
\mathtt{bab\_absent}(X), \mathtt{bba\_present}(X), \\
&&&
\mathtt{bdc\_present}(X)
\\[1.1em]

R_{14}: &
\mathtt{sev\_3}(X) &
\leftarrow &
\mathtt{bac\_absent}(X), \mathtt{bdd\_absent}(X), \\
&&&
\mathtt{baa\_absent}(X), \mathtt{baf\_absent}(X), \\
&&&
\mathtt{bab\_present}(X)
\\[1.1em]

R_{15}: &
\mathtt{sev\_2}(X) &
\leftarrow &
\mathtt{bac\_absent}(X), \mathtt{bdd\_absent}(X), \\
&&&
\mathtt{baa\_absent}(X), \mathtt{baf\_present}(X)

\end{array}
\]

\[
\begin{array}{llcl}

R_{16}: &
\mathtt{sev\_3}(X) &
\leftarrow &
\mathtt{bac\_absent}(X), \mathtt{bdd\_absent}(X), \\
&&&
\mathtt{baa\_present}(X), \mathtt{bdc\_absent}(X)
\\[1.1em]

R_{17}: &
\mathtt{sev\_4}(X) &
\leftarrow &
\mathtt{bac\_absent}(X), \mathtt{bdd\_absent}(X), \\
&&&
\mathtt{baa\_present}(X), \mathtt{bdc\_present}(X)
\\[1.1em]

R_{18}: &
\mathtt{sev\_2}(X) &
\leftarrow &
\mathtt{bac\_absent}(X), \mathtt{bdd\_present}(X)
\\[1.1em]

R_{19}: &
\mathtt{sev\_5}(X) &
\leftarrow &
\mathtt{bac\_present}(X)

\end{array}
\]
}

\section{Appendix: Normalized Confusion Matrices}
\label{app2}

\nd Table~\ref{tab:row_normalized_confusion_matricesHeat} reports the
row-normalized confusion matrices for the five methods evaluated in
Table~\ref{tab:severity_main_results}. Rows correspond to ground-truth
severity classes and columns to predicted severity classes.

\begin{table}[!t]
\caption{Row-normalized confusion matrices on the test set. Values are
percentages; diagonal entries are highlighted in bold. Each row sums to
$100\%$, subject to rounding.}
\label{tab:row_normalized_confusion_matricesHeat}
\centering
% \scriptsize has been removed to restore normal font size
\small
\setlength{\tabcolsep}{12pt} % Adds horizontal space between columns
\renewcommand{\arraystretch}{1.3} % Slightly increased vertical space between rows

\begin{minipage}[t]{0.9\textwidth}
\centering
\textbf{(a) Image-only severity baseline}\\[2pt]
\begin{tabular}{@{}c*{5}{r}@{}}
\toprule
True $\backslash$ Pred. & 1 & 2 & 3 & 4 & 5\\
\midrule
1 & \heat{72}{\textbf{72.2}} & \heat{4}{4.3} & \heat{5}{4.8} & \heat{5}{5.3} & \heat{13}{13.4}\\
2 & \heat{15}{15.1} & \heat{57}{\textbf{56.6}} & \heat{9}{9.4} & \heat{2}{1.9} & \heat{17}{17.0}\\
3 & \heat{23}{23.4} & \heat{19}{19.1} & \heat{43}{\textbf{42.6}} & \heat{6}{6.4} & \heat{9}{8.5}\\
4 & \heat{8}{7.7} & \heat{23}{23.1} & \heat{23}{23.1} & \heat{39}{\textbf{38.5}} & \heat{8}{7.7}\\
5 & \heat{25}{25.0} & \heat{25}{25.0} & \heat{25}{25.0} & \heat{0}{0.0} & \heat{25}{\textbf{25.0}}\\
\bottomrule
\end{tabular}
\end{minipage}

%\vspace{0.8em}

\begin{minipage}[t]{0.9\textwidth}
\centering
\textbf{(b) Proposed, Product (soft degree)}\\[2pt]
\begin{tabular}{@{}c*{5}{r}@{}}
\toprule
True $\backslash$ Pred. & 1 & 2 & 3 & 4 & 5\\
\midrule
1 & \heat{94}{\textbf{93.8}} & \heat{3}{2.9} & \heat{3}{3.3} & \heat{0}{0.0} & \heat{0}{0.0}\\
2 & \heat{38}{37.7} & \heat{51}{\textbf{50.9}} & \heat{11}{11.3} & \heat{0}{0.0} & \heat{0}{0.0}\\
3 & \heat{51}{51.1} & \heat{11}{10.6} & \heat{36}{\textbf{36.2}} & \heat{0}{0.0} & \heat{2}{2.1}\\
4 & \heat{46}{46.2} & \heat{8}{7.7} & \heat{31}{30.8} & \heat{8}{\textbf{7.7}} & \heat{8}{7.7}\\
5 & \heat{25}{25.0} & \heat{0}{0.0} & \heat{0}{0.0} & \heat{0}{0.0} & \heat{75}{\textbf{75.0}}\\
\bottomrule
\end{tabular}
\end{minipage}

%\vspace{0.8em}

\begin{minipage}[t]{0.9\textwidth}
\centering
\textbf{(c) Proposed, {\L}ukasiewicz (soft degree)}\\[2pt]
\begin{tabular}{@{}c*{5}{r}@{}}
\toprule
True $\backslash$ Pred. & 1 & 2 & 3 & 4 & 5\\
\midrule
1 & \heat{94}{\textbf{93.8}} & \heat{3}{2.9} & \heat{3}{3.3} & \heat{0}{0.0} & \heat{0}{0.0}\\
2 & \heat{38}{37.7} & \heat{51}{\textbf{50.9}} & \heat{11}{11.3} & \heat{0}{0.0} & \heat{0}{0.0}\\
3 & \heat{49}{48.9} & \heat{13}{12.8} & \heat{36}{\textbf{36.2}} & \heat{0}{0.0} & \heat{2}{2.1}\\
4 & \heat{46}{46.2} & \heat{8}{7.7} & \heat{23}{23.1} & \heat{8}{\textbf{7.7}} & \heat{15}{15.4}\\
5 & \heat{25}{25.0} & \heat{0}{0.0} & \heat{0}{0.0} & \heat{0}{0.0} & \heat{75}{\textbf{75.0}}\\
\bottomrule
\end{tabular}
\end{minipage}

%\vspace{0.8em}

\begin{minipage}[t]{0.9\textwidth}
\centering
\textbf{(d) Proposed, Hamacher (soft degree)}\\[2pt]
\begin{tabular}{@{}c*{5}{r}@{}}
\toprule
True $\backslash$ Pred. & 1 & 2 & 3 & 4 & 5\\
\midrule
1 & \heat{94}{\textbf{93.8}} & \heat{3}{2.9} & \heat{3}{3.3} & \heat{0}{0.0} & \heat{0}{0.0}\\
2 & \heat{38}{37.7} & \heat{51}{\textbf{50.9}} & \heat{11}{11.3} & \heat{0}{0.0} & \heat{0}{0.0}\\
3 & \heat{51}{51.1} & \heat{11}{10.6} & \heat{36}{\textbf{36.2}} & \heat{0}{0.0} & \heat{2}{2.1}\\
4 & \heat{46}{46.2} & \heat{8}{7.7} & \heat{31}{30.8} & \heat{8}{\textbf{7.7}} & \heat{8}{7.7}\\
5 & \heat{25}{25.0} & \heat{0}{0.0} & \heat{0}{0.0} & \heat{0}{0.0} & \heat{75}{\textbf{75.0}}\\
\bottomrule
\end{tabular}
\end{minipage}

%\vspace{0.8em}

\begin{minipage}[t]{0.9\textwidth}
\centering
\textbf{(e) Oracle CODE + Rule, Product}\\[2pt]
\begin{tabular}{@{}c*{5}{r}@{}}
\toprule
True $\backslash$ Pred. & 1 & 2 & 3 & 4 & 5\\
\midrule
1 & \heat{100}{\textbf{99.5}} & \heat{1}{0.5} & \heat{0}{0.0} & \heat{0}{0.0} & \heat{0}{0.0}\\
2 & \heat{28}{28.3} & \heat{57}{\textbf{56.6}} & \heat{15}{15.1} & \heat{0}{0.0} & \heat{0}{0.0}\\
3 & \heat{19}{19.1} & \heat{6}{6.4} & \heat{72}{\textbf{72.3}} & \heat{0}{0.0} & \heat{2}{2.1}\\
4 & \heat{0}{0.0} & \heat{8}{7.7} & \heat{39}{38.5} & \heat{39}{\textbf{38.5}} & \heat{15}{15.4}\\
5 & \heat{0}{0.0} & \heat{0}{0.0} & \heat{0}{0.0} & \heat{0}{0.0} & \heat{100}{\textbf{100.0}}\\
\bottomrule
\end{tabular}
\end{minipage}

\end{table}


\begin{thebibliography}{10}

\bibitem{Ana10}
EV~Ana and Willy Bauwens.
\newblock Modeling the structural deterioration of urban drainage pipes: the
  state-of-the-art in statistical methods.
\newblock {\em Urban Water Journal}, 7(1):47--59, 2010.

\bibitem{Badreddine22}
Samy Badreddine, Artur~d'Avila Garcez, Luciano Serafini, and Michael Spranger.
\newblock Logic tensor networks.
\newblock {\em Artificial Intelligence}, 303:103649, 2022.

\bibitem{Cardillo24}
Franco~Alberto Cardillo, Franca Debole, and Umberto Straccia.
\newblock Pn-owl: A two stage algorithm to learn fuzzy concept inclusions from
  owl ontologies.
\newblock {\em Fuzzy Sets and Systems}, 490(109048), 2024.

\bibitem{Cardillo22}
Franco~Alberto Cardillo and Umberto Straccia.
\newblock Fuzzy owl-boost: Learning fuzzy concept inclusions via real-valued
  boosting.
\newblock {\em Fuzzy Sets and Systems}, 438:164--186, 2022.

\bibitem{Chae01}
Myung~Jin Chae and Dulcy~M Abraham.
\newblock Neuro-fuzzy approaches for sanitary sewer pipeline condition
  assessment.
\newblock {\em Journal of Computing in Civil engineering}, 15(1):4--14, 2001.

\bibitem{Cohen95}
William~W Cohen.
\newblock Fast effective rule induction.
\newblock In {\em Proceedings of the Twelfth International Conference on
  Machine Learning (ICML)}, pages 115--123. Morgan Kaufmann, 1995.

\bibitem{Dong19}
Honghua Dong, Jiayuan Mao, Tian Lin, Chong Wang, Lihong Li, and Denny Zhou.
\newblock Neural logic machines.
\newblock {\em arXiv preprint arXiv:1904.11694}, 2019.

\bibitem{Fenner00}
R.A Fenner.
\newblock Approaches to sewer maintenance: A review.
\newblock {\em Urban Water}, 2:343--356, 12 2000.

\bibitem{Hassan19}
Syed~Ibrahim Hassan, L~Minh Dang, Irfan Mehmood, Suhyeon Im, Changho Choi,
  Jaemo Kang, Young-Soo Park, and Hyeonjoon Moon.
\newblock Underground sewer pipe condition assessment based on convolutional
  neural networks.
\newblock {\em Automation in Construction}, 106:102849, 2019.

\bibitem{Haurum22}
Joakim~Bruslund Haurum, Meysam Madadi, Sergio Escalera, and Thomas~B Moeslund.
\newblock Multi-task classification of sewer pipe defects and properties using
  a cross-task graph neural network decoder.
\newblock In {\em Proceedings of the IEEE/CVF Winter Conference on Applications
  of Computer Vision}, pages 2806--2817, 2022.

\bibitem{Haurum20}
Joakim~Bruslund Haurum and Thomas~B Moeslund.
\newblock A survey on image-based automation of cctv and sset sewer
  inspections.
\newblock {\em Automation in Construction}, 111:103061, 2020.

\bibitem{Klir87}
George~J. Klir.
\newblock Where do we stand on measures of uncertainty, ambiguity, fuzziness,
  and the like?
\newblock {\em Fuzzy Sets Syst.}, 24(2):141--160, 1987.

\bibitem{Klir95}
George~J. Klir and Bo~Yuan.
\newblock {\em Fuzzy sets and fuzzy logic: theory and applications}.
\newblock Prentice-Hall, Inc., Upper Saddle River, NJ, USA, 1995.

\bibitem{Koh20}
Pang~Wei Koh, Thao Nguyen, Yew~Siang Tang, Stephen Mussmann, Emma Pierson, Been
  Kim, and Percy Liang.
\newblock Concept bottleneck models.
\newblock In {\em International conference on machine learning}, pages
  5338--5348. PMLR, 2020.

\bibitem{Kumar18}
Srinath~S Kumar, Dulcy~M Abraham, Mohammad~R Jahanshahi, Tom Iseley, and Justin
  Starr.
\newblock Automated defect classification in sewer closed circuit television
  inspections using deep convolutional neural networks.
\newblock {\em Automation in Construction}, 91:273--283, 2018.

\bibitem{Li19}
Duanshun Li, Anran Cong, and Shuai Guo.
\newblock Sewer damage detection from imbalanced cctv inspection data using
  deep convolutional neural networks with hierarchical classification.
\newblock {\em Automation in Construction}, 101:199--208, 2019.

\bibitem{liu21}
Ze~Liu, Yutong Lin, Yue Cao, Han Hu, Yixuan Wei, Zheng Zhang, Stephen Lin, and
  Baining Guo.
\newblock Swin transformer: Hierarchical vision transformer using shifted
  windows.
\newblock In {\em Proceedings of the IEEE/CVF international conference on
  computer vision}, pages 10012--10022, 2021.

\bibitem{Meghini01}
Carlo Meghini, Fabrizio Sebastiani, and Umberto Straccia.
\newblock A model of multimedia information retrieval.
\newblock {\em Journal of the ACM}, 48(5):909--970, 2001.

\bibitem{Mitchell97}
Tom~M. Mitchell.
\newblock {\em Machine Learning}.
\newblock McGraw-Hill series in computer science. McGraw-Hill, 1 edition, 1997.

\bibitem{Quinlan87}
J.~Ross Quinlan.
\newblock Simplifying decision trees.
\newblock {\em International Journal of Man-Machine Studies}, 27(5):221--234,
  1987.

\bibitem{Quinlan93}
J.~Ross Quinlan.
\newblock {\em C4.5: Programs for Machine Learning}.
\newblock Morgan Kaufmann, 1993.

\bibitem{Sarker21}
Md~Kamruzzaman Sarker, Lu~Zhou, Aaron Eberhart, and Pascal Hitzler.
\newblock Neuro-symbolic artificial intelligence: Current trends.
\newblock {\em arXiv preprint arXiv:2105.05330}, 2021.

\bibitem{Sinha06}
Sunil~K Sinha and Paul~W Fieguth.
\newblock Neuro-fuzzy network for the classification of buried pipe defects.
\newblock {\em Automation in Construction}, 15(1):73--83, 2006.

\bibitem{Sinha02}
Sunil~K Sinha and Fakhri Karray.
\newblock Classification of underground pipe scanned images using feature
  extraction and neuro-fuzzy algorithm.
\newblock {\em IEEE Transactions on Neural Networks}, 13(2):393--401, 2002.

\bibitem{Straccia08a}
Umberto Straccia.
\newblock Managing uncertainty and vagueness in description logics, logic
  programs and description logic programs.
\newblock In {\em Reasoning Web, 4th International Summer School, Tutorial
  Lectures}, volume 5224 of {\em Lecture Notes in Computer Science}, pages
  54--103. Springer Verlag, 2008.

\bibitem{Straccia13}
Umberto Straccia.
\newblock {\em Foundations of Fuzzy Logic and Semantic Web Languages}.
\newblock CRC Studies in Informatics Series. Chapman {\&} Hall, 2013.

\bibitem{Tscheikner19}
Franz Tscheikner-Gratl, Nicolas Caradot, Fr{\'e}d{\'e}ric Cherqui, Joao~P
  Leit{\~a}o, Mehdi Ahmadi, Jeroen~G Langeveld, Yves Le~Gat, Lisa Scholten,
  Bardia Roghani, Juan~Pablo Rodr{\'\i}guez, et~al.
\newblock Sewer asset management--state of the art and research needs.
\newblock {\em Urban Water Journal}, 16(9):662--675, 2019.

\bibitem{Ullman89}
J.~D. Ullman.
\newblock {\em Principles of Database and Knowledge Base Systems}, volume 1,2.
\newblock Computer Science Press, Potomac, Maryland, 1989.

\bibitem{Wang21}
Mingzhu Wang, Han Luo, and Jack~CP Cheng.
\newblock Towards an automated condition assessment framework of underground
  sewer pipes based on closed-circuit television (cctv) images.
\newblock {\em Tunnelling and Underground Space Technology}, 110:103840, 2021.

\bibitem{Witten11}
Ian~H Witten, Eibe Frank, and Mark~A Hall.
\newblock {\em Data Mining: Practical Machine Learning Tools and Techniques}.
\newblock Morgan Kaufmann, 3rd edition, 2011.

\bibitem{Yang23}
Zhun Yang, Adam Ishay, and Joohyung Lee.
\newblock Neurasp: Embracing neural networks into answer set programming.
\newblock {\em arXiv preprint arXiv:2307.07700}, 2023.

\bibitem{Zhou22}
Qianqian Zhou, Zuxiang Situ, Shuai Teng, Hanlin Liu, Weifeng Chen, and Gongfa
  Chen.
\newblock Automatic sewer defect detection and severity quantification based on
  pixel-level semantic segmentation.
\newblock {\em Tunnelling and Underground Space Technology}, 123:104403, 2022.

\end{thebibliography}
\end{document}